\documentclass[letterpaper, journal]{IEEEtran}
\usepackage{blindtext}
\usepackage{graphicx}
\usepackage{hyperref}
\usepackage{url}

\usepackage[utf8]{inputenc} 
\usepackage[T1]{fontenc}    
\usepackage{hyperref}       
\usepackage{url}            
\usepackage{booktabs}       
\usepackage{amsfonts}       
\usepackage{amsmath}
\usepackage{amsthm}
\usepackage{nicefrac}       
\usepackage{microtype}      
\usepackage[ruled, linesnumbered]{algorithm2e}
\usepackage{subfig}
\usepackage{graphicx}
\usepackage[table]{xcolor}
\usepackage{float}
\usepackage{tabularx}
\usepackage{array}
\newcolumntype{L}[1]{>{\raggedright\let\newline\\\arraybackslash\hspace{0pt}}m{#1}}
\newcolumntype{C}[1]{>{\centering\let\newline\\\arraybackslash\hspace{0pt}}m{#1}}
\newcolumntype{R}[1]{>{\raggedleft\let\newline\\\arraybackslash\hspace{0pt}}m{#1}}
\usepackage{multirow}
\usepackage{makecell}
\usepackage{wrapfig}
\usepackage{booktabs}
\usepackage{array}
\usepackage{colortbl}
\usepackage{footnote}
\usepackage{bbm}
\usepackage{lipsum}
\usepackage{tikz}
\newcommand*\circled[1]{\tikz[baseline=(char.base)]{
            \node[shape=circle,draw,inner sep=0pt] (char) {#1};}}

\newcommand{\W}{\mathcal{W}}

\DeclareMathOperator{\msg}{msg}

\begin{document}
\title{Robust and Communication-Efficient Federated Learning from Non-IID Data}
\author{Felix Sattler, Simon Wiedemann, Klaus-Robert M{\"u}ller*,~\IEEEmembership{Member,~IEEE}, and Wojciech Samek*,~\IEEEmembership{Member,~IEEE}
\thanks{This work was supported by the Fraunhofer Society through the MPI-FhG collaboration project ``Theory \& Practice for Reduced Learning Machines''. This research was also supported by the German Ministry for Education and Research as Berlin Big Data Center (01IS14013A) and the Berlin Center for Machine Learning (01IS18037I). Partial funding by DFG is acknowledged (EXC 2046/1, project-ID: 390685689). This
work was also supported by the Information \& Communications Technology Planning \& Evaluation (IITP) grant funded by the Korea government (No. 2017-0-00451).}
\thanks{F. Sattler, S. Wiedemann and W. Samek are with Fraunhofer Heinrich Hertz Institute, 10587 Berlin, Germany (e-mail: wojciech.samek@hhi.fraunhofer.de).}
\thanks{K.-R. M{\"u}ller is with the Technische Universit{\"a}t Berlin, 10587 Berlin, Germany, with the Max Planck Institute for Informatics, 66123 Saarbr{\"u}cken, Germany, and also with the Department of Brain and Cognitive Engineering, Korea University, Seoul 136-713, South Korea (e-mail: klaus-robert.mueller@tu-berlin.de).}}
\markboth{Sattler et al.\ -- Robust and communication-efficient Federated Learning from non-iid Data}%
{Sattler et al.\ -- Robust and communication-efficient Federated Learning from non-iid Data}

\maketitle

\begin{abstract}
Federated Learning allows multiple parties to jointly train a deep learning model on their combined data, without any of the participants having to reveal their local data to a centralized server. This form of privacy-preserving collaborative learning however comes at the cost of a significant communication overhead during training. To address this problem, several compression methods have been proposed in the distributed training literature that can reduce the amount of required communication by up to three orders of magnitude. These existing methods however are only of limited utility in the Federated Learning setting, as they either only compress the upstream communication from the clients to the server (leaving the downstream communication uncompressed) or only perform well under idealized conditions such as iid distribution of the client data, which typically can not be found in Federated Learning. In this work, we propose Sparse Ternary Compression (STC), a new compression framework that is specifically designed to meet the requirements of the Federated Learning environment. STC extends the existing compression technique of top-k gradient sparsification with a novel mechanism to enable downstream compression as well as ternarization and optimal Golomb encoding of the weight updates. 
Our experiments on four different learning tasks demonstrate that STC distinctively outperforms Federated Averaging in common Federated Learning scenarios where clients either a) hold non-iid data, b) use small batch sizes during training, or where c) the number of clients is large and the participation rate in every communication round is low. 
We furthermore show that even if the clients hold iid data and use medium sized batches for training, STC still behaves pareto-superior to Federated Averaging in the sense that it achieves fixed target accuracies on our benchmarks within both fewer training iterations and a smaller communication budget. These results advocate for a paradigm shift in Federated optimization towards high-frequency low-bitwidth communication, in particular in bandwidth-constrained learning environments.       
\end{abstract}

\begin{IEEEkeywords}
Deep learning, distributed learning, Federated Learning, efficient communication, privacy-preserving machine learning.
\end{IEEEkeywords}
\IEEEpeerreviewmaketitle

\section{Introduction}
\label{sec:Intro}
Three major developments are currently transforming the ways how data is created and processed: 
First of all, with the advent of the Internet of Things (IoT), the number of intelligent devices in the world has rapidly grown in the last couple of years. Many of these devices are equipped with various sensors and increasingly potent hardware that allow them to collect and process data at unprecedented scales \cite{taylor2015world}\cite{WieArXiv18}\cite{WieArXiv18b}.

In a concurrent development deep learning has revolutionized the ways that information can be extracted from data resources with groundbreaking successes in areas such as computer vision, natural language processing or voice recognition among many others \cite{lecun2015deep}\cite{karpathy2015deep}\cite{BosTIP18}\cite{karpathy2014large}\cite{sutskever2014sequence}\cite{SamITU18b}. Deep learning scales well with growing amounts of data and it's astounding successes in recent times can be at least partly attributed to the availability of very large datasets for training. Therefore there lays huge potential in harnessing the rich data provided by IoT devices for the training and improving of deep learning models \cite{mcmahan2016communication}.

At the same time data privacy has become a growing concern for many users. Multiple cases of data leakage and misuse in recent times have demonstrated that the centralized processing of data comes at a high risk for the end users privacy.  
As IoT devices usually collect data in private environments, often even without explicit awareness of the users, these concerns hold particularly strong. 
It is therefore generally not an option to share this data with a centralized entity that could conduct training of a deep learning model. In other situations local processing of the data might be desirable for other reasons such as increased autonomy of the local agent.

This leaves us facing the following dilemma: How are we going to make use of the rich combined data of millions of IoT devices for training deep learning models if this data can not be stored at a centralized location?

\begin{figure}[t]
\centering
\includegraphics[width=0.5\textwidth]{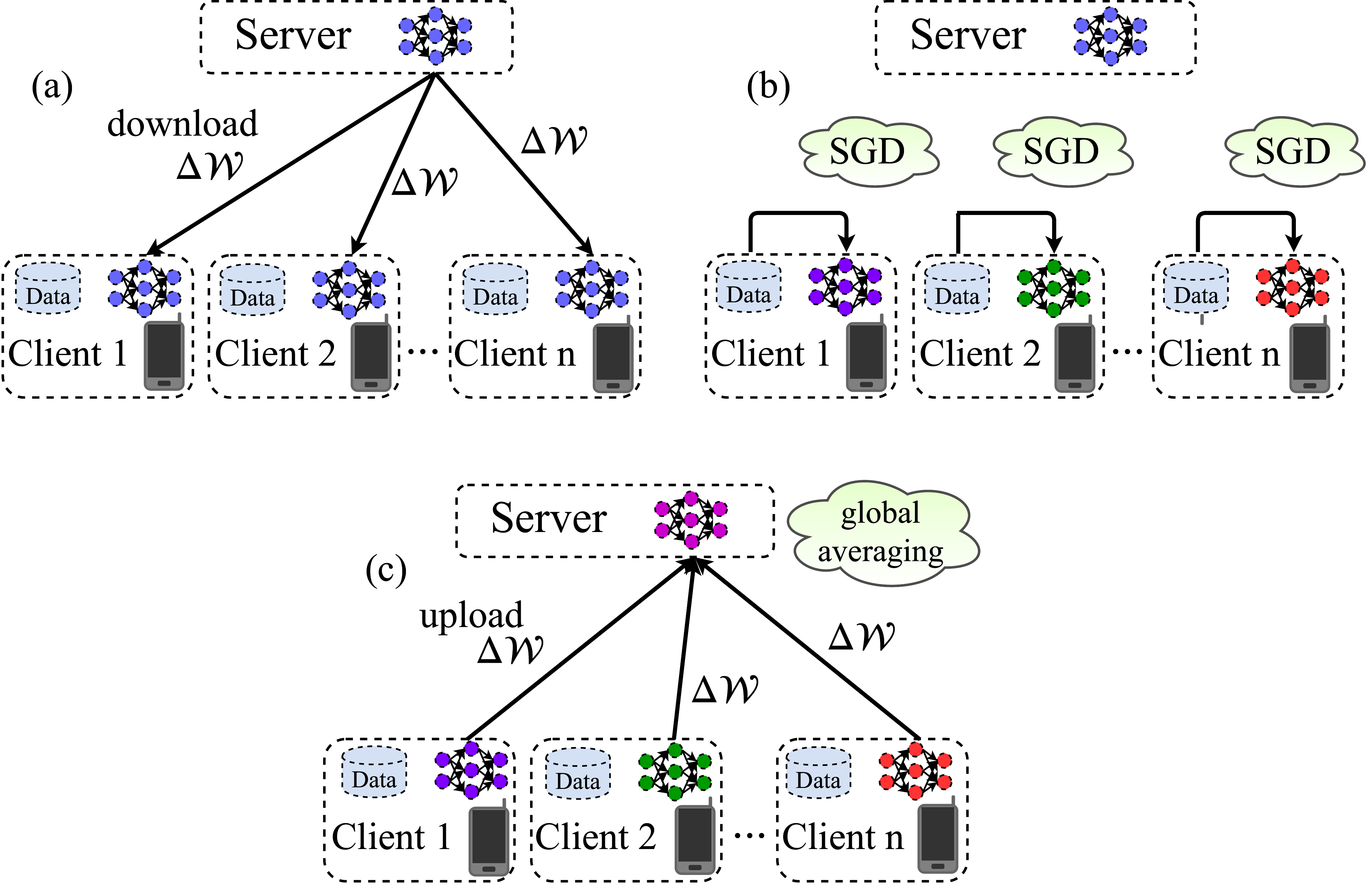}
\caption{Federated Learning with a parameter server. Illustrated is one communication round of distributed SGD: a) Clients synchronize with the server. b) Clients compute a weight update independently based on their local data.
c) Clients upload their local weight updates to the server, where they are averaged
to produce the new master model.}
\label{fig:dsgd}
\end{figure}

Federated Learning resolves this issue as it allows multiple parties to jointly train a deep learning model on their combined data, without any of the participants having to reveal their data to a centralized server \cite{mcmahan2016communication}. This form of privacy-preserving collaborative learning is achieved by following a simple three step protocol illustrated in Fig.\ \ref{fig:dsgd}. In the first step, all participating clients \emph{download} the latest master model $\W$ from the server. Next, the clients improve the downloaded model, based on their local training data using  stochastic gradient descent (SGD). Finally, all participating clients \emph{upload} their locally improved models $\W_i$ back to the server, where they are gathered and aggregated to form a new master model (in practice, weight updates $\Delta \W=\W^{new}-\W^{old}$ can be communicated instead of full models $\W$, which is equivalent as long as all clients remain synchronized). These steps are repeated until a certain convergence criterion is satisfied. Observe, that when following this protocol, training data never leaves the local devices as only model updates are communicated. Although it has been shown that in adversarial settings information about the training data can still be inferred from these updates \cite{bagdasaryan2018backdoor}, additional mechanisms such as homomorphic encryption of the updates \cite{bonawitz2017practical}\cite{hardy2017private} or differentially private training \cite{abadi2016deep} can be applied to fully conceal any information about the local data.   

A major issue in Federated Learning is the massive communication overhead that arises from sending around the model updates. When naively following the protocol described above, every participating client has to communicate a full model update during every training iteration. Every such update is of the same size as the trained model, which can be in the range of gigabytes for modern architectures with millions of parameters \cite{he2016deep}\cite{huang2017densely}. Over the course of multiple hundred thousands of training iterations on big datasets the total communication for every client can easily grow to more than a \emph{petabyte} \cite{sattler2018sparse}. Consequently, if communication bandwidth is limited or communication is costly (naive) Federated Learning can become unproductive or even completely unfeasible. 

The total amount of bits that have to be uploaded and downloaded by every client during training is given by
\begin{align}\label{eq:b_total}
\hspace*{-0.4cm}\texttt{b}^{up/down} \in \mathcal{O}( \underbrace{N_{iter}\times f}_{\text{\# updates}} \times \underbrace{|\W|\times (H(\Delta \W^{up/down})+\eta)}_{\text{update size}})
\end{align} 
where $N_{iter}$ is the total number of training iterations (forward-backward passes) performed by every client, $f$ is the communication frequency, $|\W|$ is the size of the model, $H(\Delta\W^{up/down})$ is the entropy of the weight updates exchanged during upload and download respectively, and $\eta$ is the inefficiency of the encoding, i.e.\ the difference between the true update size and the minimal update size (which is given by the entropy). If we assume the size of the model and number of training iterations to be fixed (e.g.\ because we want to achieve a certain accuracy on a given task), this leaves us with three options to reduce communication: We can a) reduce the communication frequency $f$, b) reduce the entropy of the weight updates $H(\Delta\W^{up/down})$ via lossy compression schemes and/or c) use more efficient encodings to communicate the weight updates, thus reducing $\eta$. 

\section{Challenges of the Federated Learning Environment}
\label{sec:FedLearn}
Before we can consider ways to reduce the amount of communication we first have to take into account the unique characteristics, which distinguish Federated Learning from other distributed training settings such as Parallel Training (compare also with \cite{mcmahan2016communication}). In Federated Learning the distribution of both training data and computational resources is a fundamental and fixed property of the learning environment. This entails the following challenges:

\textbf{Unbalanced and non-IID data:} As the training data present on the individual clients is collected by the clients themselves based on their local environment and usage pattern, both the size and the distribution of the local datasets will typically vary heavily between different clients.

\textbf{Large number of clients:} Federated Learning environments may constitute of multiple millions of participants \cite{bonawitz2019towards}. Furthermore, as the quality of the collaboratively learned model is determined by the combined available data of all clients, collaborative learning environments will have a natural tendency to grow. 

\textbf{Parameter server:} Once the number of clients grows beyond a certain threshold, direct communication of weight updates becomes unfeasible, because the workload for both communication and aggregation of updates grows linearly with the number of clients. In Federated Learning it is therefore unavoidable to communicate via an intermediate parameter server. This reduces the amount of communication per client and communication rounds to one single upload of a local weight update to and one download of the aggregated update from the server and moves the workload of aggregation away from the clients. Communicating via a parameter server however introduces an additional challenge to communication-efficient distributed training, as now both the upload to the server and the download from the server need to be compressed in order to reduce communication time and energy consumption. 

\textbf{Partial participation:} In the general Federated Learning for IoT setting it can generally not be guaranteed that all clients participate in every communication round. Devices might loose their connection, run out of battery or seize to contribute to the collaborative training for other reasons.   

\textbf{Limited battery and memory:}
Mobile and embedded devices often are not connected to a power grid. Instead their capacity to run computations is limited by a finite battery. Performing iterations of stochastic gradient descent is notoriously expensive for deep neural networks. It is therefore necessary to keep the number of gradient evaluations per client as small as possible. Mobile and embedded devices also typically have only very limited memory. As the memory footprint of SGD grows linearly with the batch size, this might force the devices to train on very small batch sizes. 

\ \\
Based on the above characterization of the Federated Learning environment we conclude that a communication-efficient distributed training algorithm for Federated Learning needs to fulfill the following requirements:
\begin{itemize}
\item[\textbf{(R1)}] It should compress both upstream and downstream communication. 
\item[\textbf{(R2)}] It should be robust to non-iid, small batch sizes and unbalanced data.
\item[\textbf{(R3)}] It should be robust to large numbers of clients and partial client participation.
\end{itemize}

\ \\
In this work we will demonstrate that none of the existing methods proposed for communication-efficient Federated Learning satisfies \emph{all} of these requirements (cf.\ Table \ref{tab:compare}). More concretely we will show, that the methods which are able to compress both upstream and downstream communication are very sensitive to non-iid data distributions, while the methods which are more robust to this type of data do not compress the downstream (Section \ref{sec:distributedtraining}). We will then proceed to construct a new communication protocol that resolves these issues and meets all requirements \textbf{(R1)} - \textbf{(R3)}. We will provide extensive empirical results on four different neural network architectures and datasets that will demonstrate that our protocol is superior to existing compression schemes in that it requires both fewer gradient evaluations and communicated bits to converge to a given target accuracy (Section \ref{sec:conclusion}). These results also extend to the iid regime. 

\section{Related Work} 
\label{sec:relatedwork}

\begin{table}[t]
\caption{Different methods for communication-efficient distributed deep learning proposed in the literature. None of the existing methods satisfies all requirements \textbf{(R1)} - \textbf{(R3)} of the Federated Learning environment. We call a method "robust to non-iid data" if the federated training converges independent of the local distribution of client data. We call compression rates greater than $\times 32$ "strong" and those smaller or equal to $\times 32$ "weak".}
\label{tab:compare}
\centering
\scalebox{0.5}{
\Large

\setlength\tabcolsep{3pt}

{\renewcommand{\arraystretch}{1.8}
\begin{tabular}{|c|c|c|c|}
\hline
\textbf{Method} & \makecell{\textbf{Downstream}\\\textbf{Compression}} & \makecell{\textbf{Compression}\\\textbf{Rate}} & \makecell{\textbf{Robust to}\\\textbf{NON-IID Data}}\\
\hline
\hline
\makecell{TernGrad \cite{wen2017terngrad}, QSGD \cite{alistarh2017qsgd},\\ ATOMO \cite{wang2018atomo}} &  NO & WEAK & NO \\
 \hline
 signSGD \cite{bernstein2018signsgd}& \cellcolor{green!25}YES & WEAK & NO\\
\hline
\makecell{Gradient Dropping \cite{aji2017sparse}, DGC \cite{lin2017deep},\\ Variance based \cite{strom2015scalable}, Strom \cite{tsuzuku2018variance}}& NO & \cellcolor{green!25}STRONG & \cellcolor{green!25}YES \\
\hline
Federated Averaging \cite{mcmahan2016communication} & \cellcolor{green!25}YES & \cellcolor{green!25} STRONG & NO\\
\hline
\hline
\makecell{Sparse Ternary\\Compression (ours)} & \cellcolor{green!25}YES & \cellcolor{green!25} STRONG & \cellcolor{green!25} YES\\
\hline
\end{tabular}
}
}

\end{table}

In the broader realm of communication-efficient distributed deep learning, a wide variety of methods has been proposed to reduce the amount of communication during the training process. 
Using equation \eqref{eq:b_total} as a reference, we can organize the substantial existing research body on communication-efficient distributed deep learning into three different groups:

\textbf{Communication delay} methods reduce the communication frequency $f$. McMahan et al.\ \cite{mcmahan2016communication} propose Federated Averaging where instead of communicating after every iteration, every client performs multiple iterations of SGD to compute a weight update. The authors observe that on different convolutional and recurrent neural network architectures communication can be delayed for up to 100 iterations without significantly affecting the convergence speed as long as the data is distributed among the clients in an iid manner. The amount of communication can be reduced even further with longer delay periods, however this comes at the cost of an increased number of gradient evaluations. 
In a follow-up work Konecny et al. \cite{konevcny2016federated} combine this communication delay with random sparsification and probabilistic quantization. They restrict the clients to learn random sparse weight updates or force random sparsity on them afterwards ("structured" vs "sketched" updates) and combine this sparsification with probabilistic quantization. Their method however significantly slows down convergence speed in terms of SGD iterations.
Communication delay methods automatically reduce both upstream and downstream communication and are proven to work with large numbers of clients and partial client participation.

\textbf{Sparsification} methods reduce the entropy $H(\Delta \W)$ of the updates by restricting changes to only a small subset of the parameters. Strom \cite{strom2015scalable} presents an approach (later modified by \cite{tsuzuku2018variance}) in which only gradients with a magnitude greater than a certain predefined threshold are sent to the server. All other gradients are accumulated in a residual. This method is shown to achieve upstream compression rates of up to 3 orders of magnitude on an acoustic modeling task. In practice however, it is hard to choose appropriate values for the threshold, as it may vary a lot for different architectures and even different layers. 
To overcome this issue Aji et al. \cite{aji2017sparse} instead fix the sparsity \emph{rate} and only communicate the fraction $p$ entries with the biggest magnitude of each gradient, while also collecting all other gradients in a residual. At a sparsity rate of $p=0.001$ their method only slightly degrades the convergence speed and final accuracy of the trained model. Lin et al. \cite{lin2017deep} present minor modifications to the work of Aji et al. which even close this small performance gap. Sparsification methods have been proposed primarily with the intention to speed up parallel training in the data center. Their convergence properties in the much more challenging Federated Learning environments have not yet been investigated. Sparsification methods (in their existing form) primarily compress the upstream communication, as the sparsity patterns on the updates from different clients will generally differ. If the number of participating clients is greater than the inverse sparsity rate, which can easily be the case in Federated Learning, the downstream update will not even be compressed at all. 

\textbf{Dense quantization} methods reduce the entropy of the weight updates by restricting all updates to a reduced set of values. Bernstein et al.\ propose signSGD \cite{bernstein2018signsgd}, a compression method with theoretical convergence guarantees on iid data that quantizes every gradient update to it's binary sign, thus reducing the bit size per update by a factor of $\times32$. signSGD also incorporates download compression by aggregating the binary updates from all clients by means of a majority vote.
Other authors propose to stochastically quantize the gradients during upload in an unbiased way (TernGrad \cite{wen2017terngrad}, QSGD \cite{alistarh2017qsgd}, ATOMO \cite{wang2018atomo}). These methods are theoretically appealing, as they inherit the convergence properties of regular SGD under relatively mild assumptions. However their empirical performance and compression rates do not match those of sparsification methods. 


Out of all the above listed methods, only \emph{Federated Averaging} and \emph{signSGD} compress both the upstream and downstream communication. All other methods are of limited utility in the Federated Learning setting defined in Section \ref{sec:FedLearn} as they leave the communication from the server to the clients uncompressed.

\ \\
\textbf{Notation:}
In the following calligraphic $\W$ will refer to the entirety of parameters of a neural network, while regular uppercase $W$ refers to one specific tensor of parameters within $\W$ and lowercase $w$ refers to one single scalar parameter of the network. Arithmetic operations between neural network parameters are to be understood element-wise. 

\section{Limitations of existing compression methods}
\label{sec:distributedtraining}

\begin{figure}[t]
\centering
\includegraphics[width=0.5\textwidth]{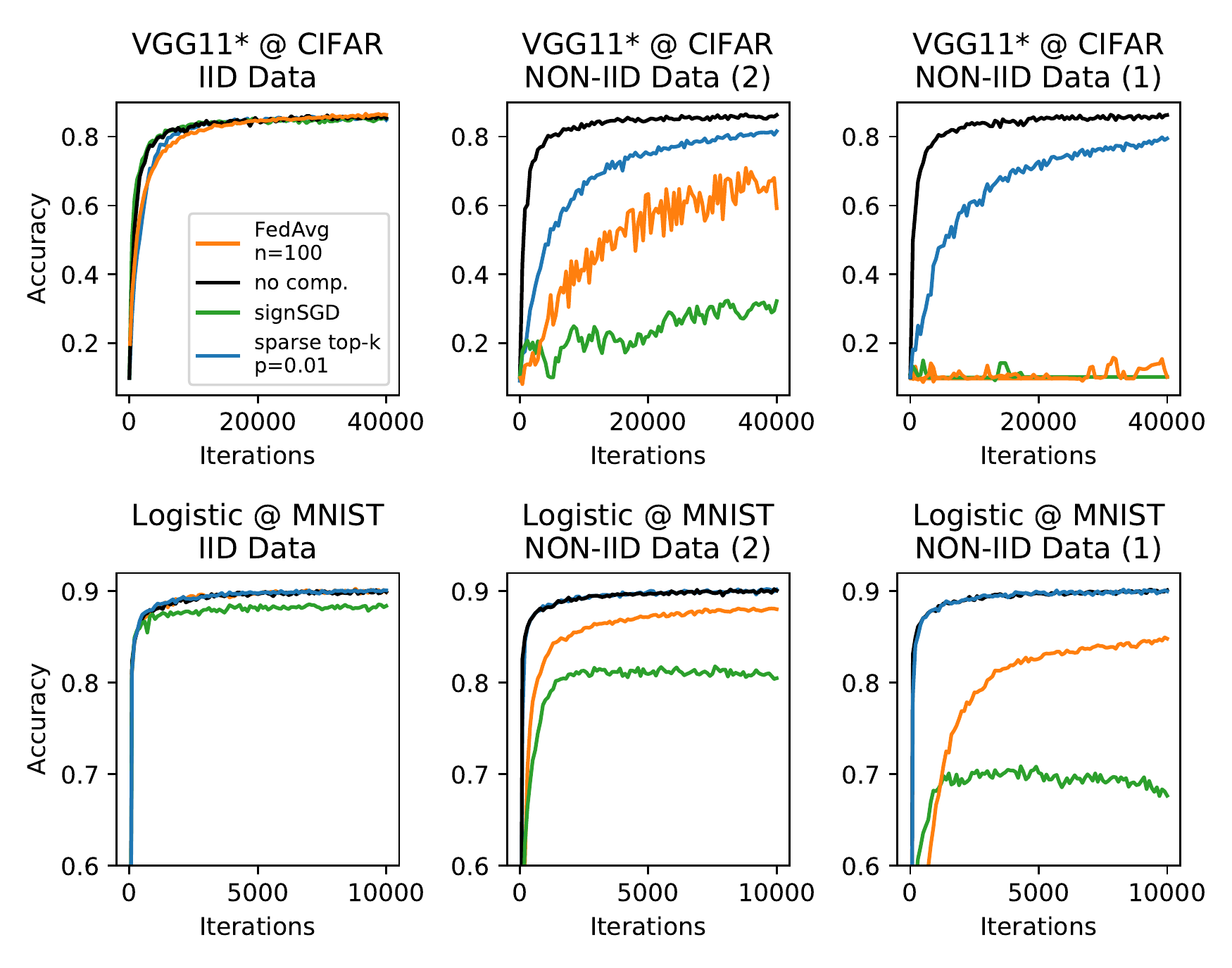}
\caption{Convergence speed when using different compression methods during the training of VGG11*\protect\footnotemark on CIFAR-10 and Logistic Regression on MNIST and Fashion-MNIST in a distributed setting with 10 clients for iid and non-iid data. In the non-iid cases, every client only holds examples from exactly two respectively one of the 10 classes in the dataset. All compression methods suffer from degraded convergence speed in the non-iid situation, but sparse top-k is affected by far the least.}
\label{fig:compare_all_iid_noniid}
\end{figure}
\footnotetext{We denote by VGG11* a simplified version of the original VGG11 architecture described in \cite{simonyan2014very}, where all dropout and batch normalization layers are removed and the number of convolutional filters and size of all fully-connected layers is reduced by a factor of 2.}

The related work on efficient distributed deep learning almost exclusively considers iid data distributions among the clients, i.e.\ they assume unbiasedness of the local gradients with respect to the full-batch gradient according to
\begin{align}
\mathbb{E}_{x\sim p_i}[\nabla_\W l(x, \W)] = \nabla_\W R(\W)~\forall i=1,..,n
\label{eq:assumpt}
\end{align} 
where $p_i$ is the distribution of data on the $i$-th client and $R(\W)$ is the empirical risk function over the combined training data. 

While this assumption is reasonable for parallel training where the distribution of data among the clients is chosen by the practitioner, it is typically not valid in the Federated Learning setting where we can generally only hope for unbiasedness in the mean
\begin{align}
\label{eq:mean_ub}
\frac{1}{n}\sum_{i=1}^n\mathbb{E}_{x^i\sim p_i}[\nabla_\W l(x^i,\W)] = \nabla_\W R(\W)
\end{align}
while the individual client's gradients will be biased towards the local dataset according to
\begin{align}
\mathbb{E}_{x\sim p_i}[\nabla_\W l(x,\W)] = \nabla_\W R_i(\W) \neq \nabla_\W R(\W)~\forall i=1,..,n.
\end{align}

As it violates assumption \eqref{eq:assumpt}, a non-iid distribution of the local data renders existing convergence guarantees as formulated in \cite{wen2017terngrad}\cite{alistarh2017qsgd}\cite{bernstein2018signsgd2}\cite{wang2018atomo} inapplicable and has dramatic effects on the practical performance of communication-efficient distributed training algorithms as we will demonstrate in the following experiments. 

\subsection{Preliminary Experiments}
We run preliminary experiments with a simplified version of the well-studied 11-layer VGG11 network \cite{simonyan2014very}, which we train on the CIFAR-10 \cite{krizhevsky2014cifar} dataset in a Federated Learning setup using 10 clients. For the iid setting we split the training data randomly into equally sized shards and assign one shard to every one of the clients. For the "non-iid ($m$)" setting we assign every client samples from exactly $m$ classes of the dataset. The data splits are non-overlapping and balanced such that every client ends up with the same number of data points.
The detailed procedure that generates the split of data is described in Section \ref{sec:split_data} of the appendix. 
We also perform experiments with a simple logistic regression classifier, which we train on the MNIST dataset \cite{lecun1998mnist} under the same setup of the Federated Learning environment. 
Both models are trained using momentum SGD. To make the results comparable, all compression methods use the same learning rate and batch size.

\subsection{Results} 
Figure \ref{fig:compare_all_iid_noniid} shows the convergence speed in terms of gradient evaluations for the two models when trained using different methods for communication-efficient Federated Learning. We observe that while all compression methods achieve comparably fast convergence in terms of gradient evaluations on iid data, closely matching the uncompressed baseline (black line), they suffer considerably in the non-iid training settings. As this trend can be observed also for the logistic regression model we can conclude that the underlying phenomenon is not unique to deep neural networks and also carries over to convex objectives. We will now analyze these results in detail for the different compression methods.

\textbf{Federated Averaging:}
Most noticeably, Federated Averaging \cite{mcmahan2016communication} (orange line in Fig.\ \ref{fig:compare_all_iid_noniid}), although specifically proposed for the Federated Learning setting, suffers considerably from non-iid data. This observation is consistent with Zhao et al.\ \cite{zhao2018federated} who demonstrated that model accuracy can drop by up to 55\% in non-iid learning environments as compared to iid ones. They attribute the loss in accuracy to the increased weight divergence between the clients and propose to side-step the problem by assigning a shared public iid dataset to all clients. While this approach can indeed create more accurate models it also has multiple shortcomings, the most crucial one being that we generally can not assume the availability of such a public dataset. If a public dataset were to exist one could use it to pre-train a model at the server, which is not consistent with the assumptions typically made in Federated Learning. Furthermore, if all clients share (part of) the same public dataset, overfitting to this shared data can become a serious issue. This effect will be particularly severe in highly distributed settings where the number of data points on every client is small. Lastly, even when sharing a relatively large dataset between the clients, the original accuracy achieved in the iid situation can not be fully restored. For these reasons, we believe that the data sharing strategy proposed by \cite{zhao2018federated} is an insufficient workaround to the fundamental problem of Federated Averaging having convergence issues on non-iid data. 

\textbf{SignSGD:}
The quantization method signSGD \cite{bernstein2018signsgd2} (green line in Fig.\ \ref{fig:compare_all_iid_noniid}) suffers from even worse stability issues in the non-iid learning environment. The method completely fails to converge on the CIFAR benchmark and even for the convex logistic regression objective the training plateaus at a substantially degraded accuracy.

To understand the reasons for these convergence issues we have to investigate how likely it is for a single batch-gradient to have the "correct" sign. 
Let 
\begin{align}
g^k_w=\frac{1}{k}\sum_{i=1}^k\nabla_w l(x_i,\W)
\end{align}
be the batch-gradient over a specific mini-batch of data $D^k=\{x_1,..,x_k\}\subset D$ of size $k$ at parameter $w$. Let further $g_w$ be the gradient over the entire training data $D$. Then  we can define this probability by
\begin{align}
\alpha_w(k)=\mathbb{P}[\text{sign}(g^k_w)=\text{sign}(g_w)].
\end{align}
We can also compute the mean statistic 
\begin{align}
\alpha(k) = \frac{1}{|\W|}\sum_{w\in\W}\alpha_w(k)
\end{align}
to estimate the average congruence over all parameters of the network.

\begin{figure}[]
\centering
\includegraphics[width=0.4\textwidth]{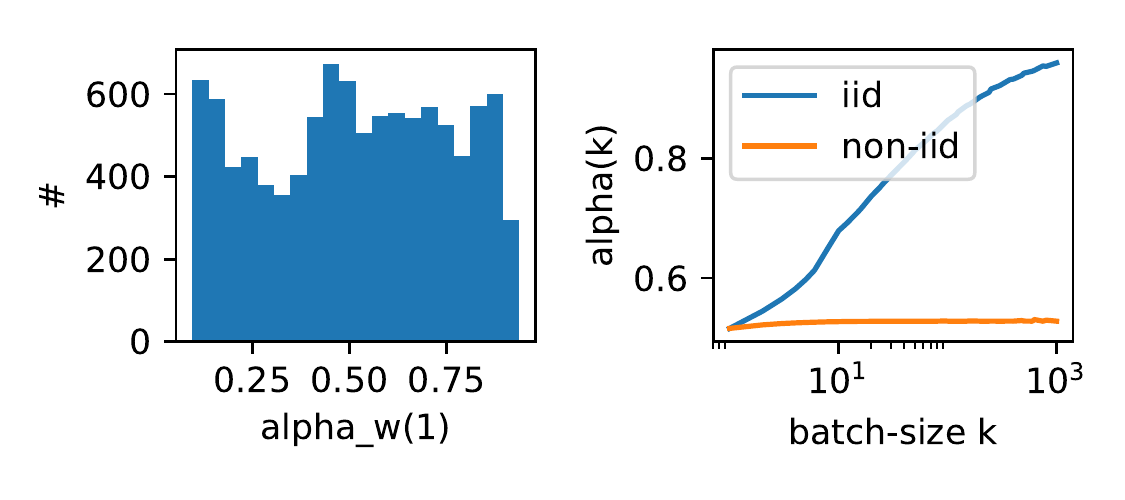}
\caption{\emph{Left}: Distribution of values for $\alpha_w(1)$ for the weight layer of a logistic regression over the MNIST dataset. \emph{Right}: Development of $\alpha(k)$ for increasing batch sizes. In the iid case the batches are sampled randomly from the training data, while in the non-iid case every batch contains samples from only exactly one class. For iid batches the gradient sign becomes increasingly accurate with growing batch sizes. For non-iid batches of data this is not the case. The gradient signs remain highly incongruent with the full-batch gradient, no matter how large the size of the batch.}
\label{fig:alpha}
\end{figure}

Figure \ref{fig:alpha} (left) exemplary shows the distribution of values for $\alpha_w(1)$ within the weights of a logistic regression on MNIST at the beginning of training. As we can see, at a batch size of 1, $g^1_w$ is a very bad predictor of the true gradient sign with a very high variance and an average congruence of $\alpha(1) = 0.51$ just slightly higher than random.
The sensitivity of signSGD to non-iid data becomes apparent once we inspect the development of the gradient sign congruence for increasing batch sizes. Figure \ref{fig:alpha} (right) shows this development for batches of increasing size sampled from an iid and non-iid distribution. For the latter one every sampled batch only contains data from exactly one class. 
As we can see, for iid data $\alpha$ quickly grows with increasing batch size, resulting in increasingly accurate updates. For non-iid data however the congruence stays low, independent of the size of the batch. This means that if clients hold highly non-iid subsets of data, signSGD updates will only weakly correlate with the direction of steepest descent, no matter how large of a batch size is chosen for training.

\textbf{Top-k Sparsification:}
Out of all existing compression methods, top-k sparsification (blue line in Fig.\ \ref{fig:compare_all_iid_noniid}) suffers least from non-iid data. For VGG11 on CIFAR the training still converges reliably, even if every client only holds data from exactly one class and for the logistic regression classifier trained on MNIST the convergence does not slow down at all. We hypothesize that this robustness to non-iid data is due to mainly two reasons: First of all, the frequent communication of weight updates between the clients prevents them from diverging too far from one another and hence top-k sparsification does not suffer from weight divergence \cite{zhao2018federated} as it is the case for Federated Averaging. Second, sparsification does not destabilize the training nearly as much as signSGD does since the noise in the stochastic gradients is not amplified by quantization. 
Although top-k sparsification shows promising performance on non-iid data, it's utility is limited in the Federated Learning setting as it only directly compresses the upstream communication.

Table \ref{tab:compare} summarizes our findings: None of the existing compression methods supports both download compression and properly works with non-iid data.

\section{Sparse Ternary Compression}
\label{sec:stc}

Top-k sparsification shows the most promising performance in distributed learning environments with non-iid client data. We will use this observation as a starting point to construct an efficient communication protocol for Federated Learning. To arrive at this protocol we have to solve two open problems, which prevent the direct application of top-k sparsification to Federated Learning:
\begin{itemize}
\item We will incorporate downstream compression into the method to allow for efficient communication from server to clients.
\item We will implement a caching mechanism to keep the clients synchronized in case of partial client participation.
\end{itemize}
Finally, we will also further increase the efficiency of our method by employing quantization and optimal lossless coding of the weight updates.

\subsection{Extending to Downstream Compression}
\label{sec:down}
Let $\text{top}_{p\%}: \mathbb{R}^n\rightarrow\mathbb{R}^n, \Delta\W\mapsto\tilde{\Delta\W}$ be the compression operator that maps a (flattened) weight update $\Delta\W$ to a sparsified weight update $\tilde{\Delta \W}$ by setting all but the fraction $p\%$ elements with the highest magnitude to zero.
For local weight updates $\Delta\W_i^{(t)}$ the update rule for top-k sparsified communication as proposed in \cite{lin2017deep} and \cite{aji2017sparse} can then be written as
\begin{align}
\label{eq:update_up}
\Delta \W^{(t+1)} &= \frac{1}{n}\sum_{i=1}^n\underbrace{\text{top}_{p\%}(\Delta \W_i^{(t+1)}+A_i^{(t)})}_{\tilde{\Delta \W_i}^{(t+1)}},\\
A_i^{(t+1)} &= A_i^{(t)}+\Delta\W_i^{(t+1)}-\tilde{\Delta \W_i}^{(t+1)},
\end{align}
starting with an empty residual $A_i^{(0)}=0\in\mathbb{R}^n$ on all clients.
While the updates $\tilde{\Delta \W_i^{(t+1)}}$ that are sent from clients to server are always sparse, the number of non-zero elements in the update $\Delta \W^{(t+1)}$ that is sent downstream grows linearly with the amount of participating clients in the worst case. If the participation rate exceeds the inverse sparsity $1/p$, the update $\Delta \W^{(t+1)}$ essentially becomes dense. 
 
To resolve this issue, we propose to apply the same compression mechanism at the server side to compress the downstream communication. This modifies the update-rule to
\begin{align}
\label{eq:update_updown}
\hspace*{-0.5cm}\tilde{\Delta \W^{(t+1)}} &= \text{top}_{p\%}(\frac{1}{n}\sum_{i=1}^n\underbrace{\text{top}_{p\%}(\Delta \W_i^{(t+1)}+A_i^{(t)})}_{\tilde{\Delta \W_i}^{(t+1)}}+A^{(t)})
\end{align}
with a client-side and a server-side residual update
\begin{align}
A_i^{(t+1)} &= A_i^{(t)}+\Delta\W_i^{(t+1)}-\tilde{\Delta \W_i}^{(t+1)}\\
A^{(t+1)} &= A^{(t)}+\Delta\W^{(t+1)}-\tilde{\Delta \W}^{(t+1)}.
\end{align}

\begin{figure}[t]
\centering
\includegraphics[width=0.5\textwidth]{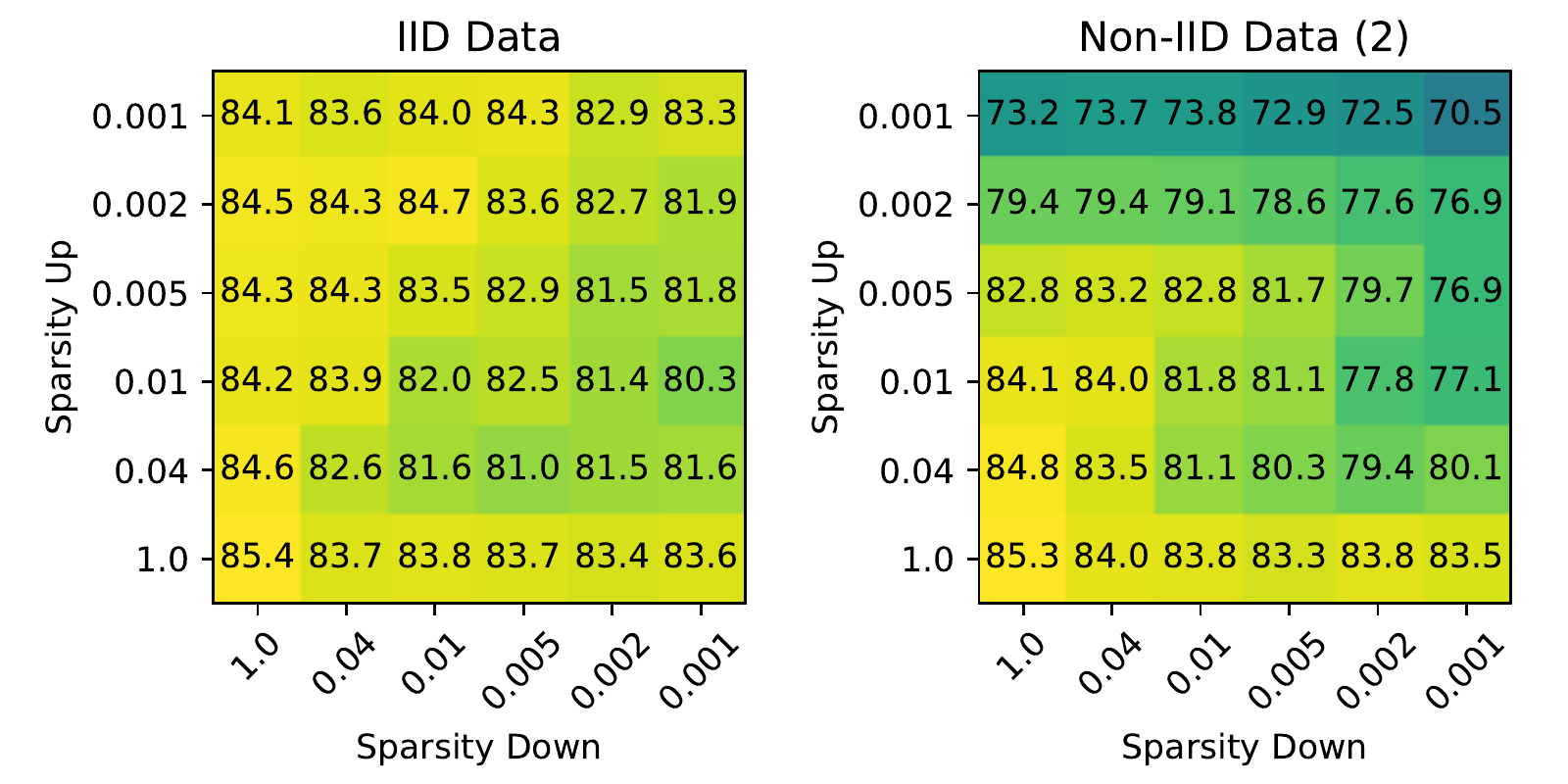}
\caption{Accuracy achieved by VGG11* when trained on CIFAR in a distributed setting with 5 clients for 16000 iterations at different levels of upload and download sparsity. Sparsifying the updates for downstream communication reduces the final accuracy by at most 3\% when compared to using only upload sparsity.}
\label{fig:compare_dgc_updown}
\end{figure}

We can express this new update rule for both upload and download compression \eqref{eq:update_updown} as a special case of pure upload compression \eqref{eq:update_up} with generalized filter masks: Let $M_i$, $i=1,..,n$ be the sparsifying filter masks used by the respective clients during the upload and $M$ be the one used during the download by the server. Then we could arrive at the same sparse update $\tilde{\Delta \W}^{(t+1)}$ if all clients use filter masks $\tilde{M}_i=M_i\odot M$, where $\odot$ is the Hadamard product.
We can thus predict that training models using this new update rule should behave similar to regular top-k sparsification with an increased sparsity rate. We can easily verify this prediction: 

Figure \ref{fig:compare_dgc_updown} shows the accuracies achieved by VGG11 on CIFAR10, when trained in a Federated Learning environment with 5 clients for 10000 iterations at different rates of upload and download compression. As we can see, for as long as download and upload sparsity are of the same order, sparsifying the download is not very harmful to the convergence and decreases the accuracy by at most two percent in both the iid and the non-iid case. 

\subsection{Weight Update Caching for Partial Client Participation}
\label{sec:part}

This far we have only been looking at scenarios in which all of the clients participate throughout the entire training process. However, as elaborated in Section \ref{sec:FedLearn}, in Federated Learning typically only a fraction of the entire client population will participate in any particular communication round. As clients do not download the full model $\W^{(t)}$, but only compressed model updates $\Delta\tilde{\W}^{(t)}$, this introduces new challenges when it comes to keeping all clients synchronized. 

To solve the synchronization problem and reduce the workload for the clients we propose to use a caching mechanism on the server. Assume the last $\tau$ communication rounds have produced the updates $\{\tilde{\Delta\W}^{(t)}|t=T-1,..,T-\tau\}$. The server can cache all partial sums of these updates up until a certain point $\{P^{(s)}=\sum_{t=1}^s \tilde{\Delta\W}^{(T-t)}|s=1,..,\tau\}$ together with the global model $\W^{(T)}=\W^{(T-\tau-1)}+\sum_{t=1}^\tau\tilde{\Delta\W}^{(T-t)}$. Every client that wants to participate in the next communication round then has to first synchronize itself with the server by either downloading $P^{(s)}$ or $\W^{(T)}$, depending on how many previous communication rounds it has skipped.
For general sparse updates the bound on the entropy
\begin{align}
H(P^{(\tau)}) \leq \tau H(P^{(1)})=\tau H(\tilde{\Delta\W}^{(T-1)})
\end{align}
can be attained. This means that the size of the download will grow linearly with the amount of rounds a client has skipped training. The average number of skipped rounds is equal to the inverse participation fraction $1/\eta$. This is usually tolerable as the down-link typically is cheaper and has far higher bandwidth than the up-link as already noted in \cite{mcmahan2016communication} and \cite{wen2017terngrad}. Essentially all compression methods, that communicate only parameter updates instead of full models suffer from this same problem. This is also the case for signSGD, although here the size of the downstream update only grows logarithmically with the delay period according to
\begin{align}
H(P_{signSGD}^{(\tau)})\leq\log_2(2\tau+1).
\end{align}

Partial client participation also has effects on the convergence speed of Federated training, both with delayed and sparsified updates. We will investigate these effects in detail in Section \ref{sec:learning_env}.

\subsection{Eliminating Redundancy}
In the two previous Sections \ref{sec:down} and \ref{sec:part} we have established that sparsified communication can be seamlessly integrated into Federated Learning. We will now look at ways to further improve the efficiency of our method, by eliminating the remaining sources of redundancy in the communication.

\textbf{Combining Sparsity with Binarization:}
Regular top-k sparsification as proposed in \cite{aji2017sparse} and \cite{lin2017deep} communicates the fraction of largest elements at full precision, while all other elements are not communicated at all. In our previous work (Sattler et al. \cite{sattler2018sparse}) we already demonstrated that this imbalance in update precision is wasteful and that higher compression gains can be achieved when sparsification is combined with quantization of the non-zero elements.  

We adopt the method described in \cite{sattler2018sparse} and quantize the remaining top-k elements of the sparsified updates to the mean population magnitude, leaving us with a ternary tensor containing values $\{-\mu,0,\mu\}$. The quantization method is formalized in Algorithm \ref{alg:STC}.

\begin{figure}
\centering
\includegraphics[width=0.5\textwidth]{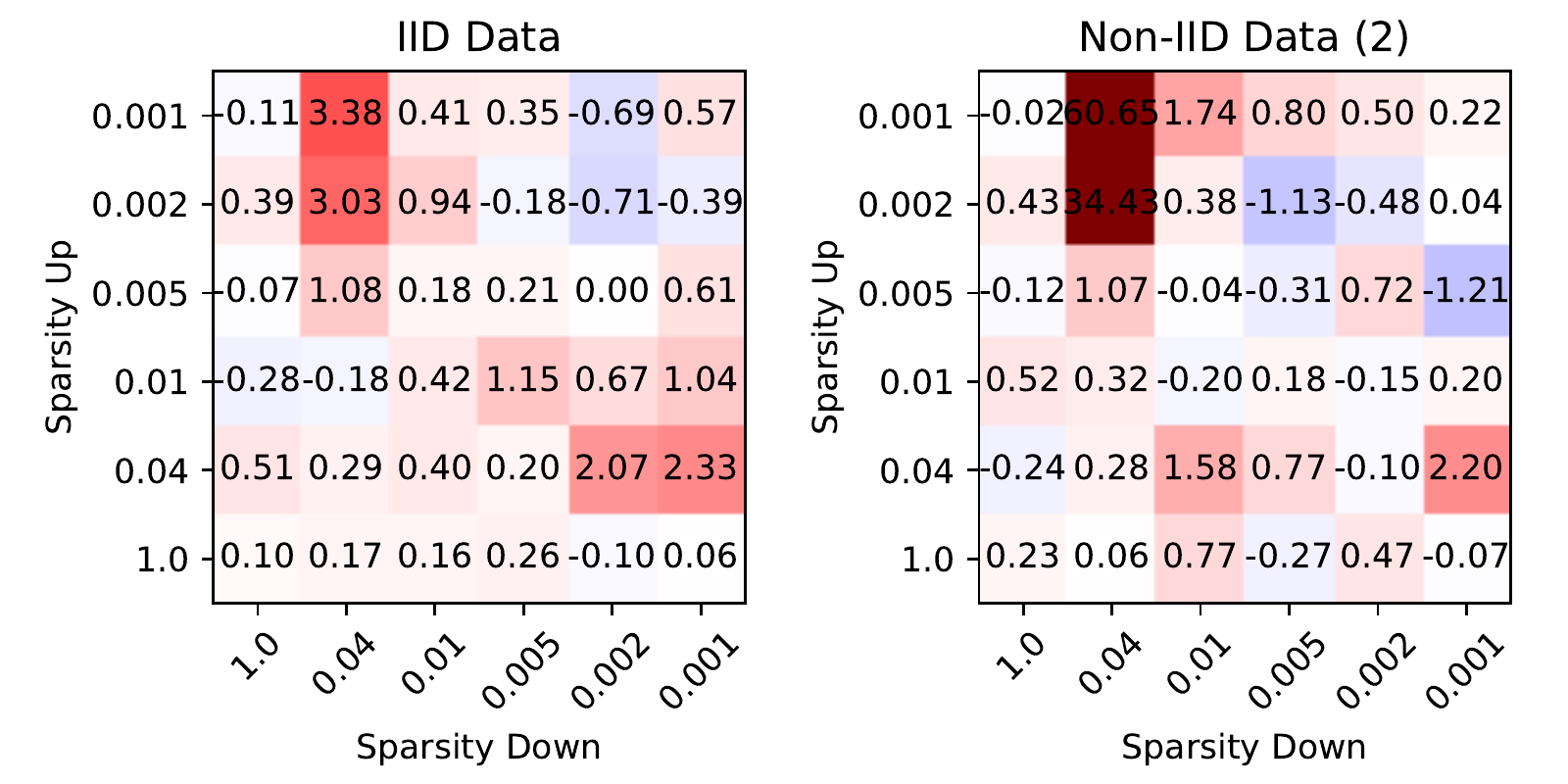}
\caption{The effects of binarization at different levels of upload- and download sparsity. Displayed is the difference in final accuracy in \% between a model trained with sparse updates and a model trained with sparse binarized updates. Positive numbers indicate better performance of the model trained with pure sparsity. VGG11 trained on CIFAR10 for 16000 iterations with 5 clients holding iid and non-iid data.}
\label{fig:compare_dgc_stc}
\end{figure}

This ternarization step reduces the entropy of the update from
\begin{align}
H_{sparse} = -p\log_2(p)-(1-p)\log_2(p)+32p
\end{align} 
to
\begin{align}
H_{STC} = -p\log_2(p)-(1-p)\log_2(p)+p
\end{align} 
when compared to regular sparsification. 
At a sparsity rate of $p=0.01$, the additional compression achieved by ternarization is $H_{sparse}/H_{STC} = 4.414$.
In order to achieve the same compression gains by pure sparsification one would have to increase the sparsity rate by approximately the same factor. Figure \ref{fig:compare_dgc_stc} shows the final accuracy of the VGG11* model when trained at different sparsity levels with and without ternarization. As we can see, additional ternarization does only have a very minor effect on the convergence speed and sometimes does even increase the final accuracy of the trained model. It seems evident that a combination of sparsity and quantization makes more efficient use of the communication budged than pure sparsification. We therefore make use of ternarization in the weight update compression of both the clients and the server.

\begin{algorithm} 
\caption{Sparse Ternary Compression (STC)}\label{alg:STC}
\DontPrintSemicolon
\textbf{input:} flattened tensor $T\in\mathbb{R}^n$, sparsity $p$\\
\textbf{output:} sparse ternary tensor $T^*\in\{-\mu,0,\mu\}^n$\\
\textbullet~ $k \leftarrow \max(np,1)$\\
\textbullet~ $v \leftarrow \text{top}_k(|T|)$\\
\textbullet~ $\text{mask} \leftarrow (|T|\geq v) \in \{0,1\}^n$\\
\textbullet~ $T^{masked}\leftarrow \text{mask}\odot T$\\
\textbullet~ $\mu \leftarrow \frac{1}{k}\sum_{i=1}^n|T^{masked}_i|$\\
\Return $T^* \leftarrow \mu\times\text{sign}(T^{masked})$\\
\end{algorithm}

\textbf{Lossless Encoding:}
To communicate a set of sparse ternary tensors produced by the above described compression scheme, we only need to transfer the positions of the non-zero elements in the flattened tensors, along with one bit per non-zero update to indicate the mean sign $\mu$ or $-\mu$. Instead of communicating the absolute positions of the non-zero elements it is favorable to communicate the distances between them. Assuming a random sparsity pattern we know that for
big values of $|W|$ and $k=p|W|$, the distances are approximately geometrically distributed with success
probability equal to the sparsity rate $p$. Therefore, we can optimally encode the distances using the Golomb code \cite{golomb1966run}. Golomb encoding reduces the average number of position bits to 
\begin{align}
\bar{\texttt{b}}_{pos} = \mathbf{b}^*+\frac{1}{1-(1-p)^{2^{\mathbf{b}^*}}},
\end{align}
with $\mathbf{b}^*=1+\lfloor \log_2(\frac{\log(\phi-1)}{\log(1-p)})\rfloor$ and $\phi=\frac{\sqrt{5}+1}{2}$ being the golden ratio. For a sparsity rate of e.g.\ $p=0.01$, we get $\bar{\texttt{b}}_{pos}=8.38$, which translates to $\times 1.9$ compression, compared to a naive distance encoding with 16 fixed bits.  Both the encoding and the decoding scheme can be found in Section \ref{sec:coding} of the appendix (Algorithms \ref{alg:encode} and \ref{alg:decode}). The updates are encoded both before upload and before download.

\ \\
The complete compression framework that features upstream and downstream compression via sparsification, ternarization and optimal encoding of the updates is described in Algorithm \ref{alg:DSGD}.

\begin{algorithm} 
\caption{Efficient Federated Learning with Parameter Server via Sparse Ternary Compression}\label{alg:DSGD}
\DontPrintSemicolon
\textbf{input:} initial parameters $\W$\\
\textbf{output:} improved parameters $\W$\\
\textbf{init:} all clients $C_i$, $i=1,..,\textnormal{[Number of Clients]}$ are initialized with the same parameters $\W_i\leftarrow \W$. Every Client holds a different dataset $D_i$, with $|\{y:(x,y)\in D_i\}|=\textnormal{[Classes per Client]}$ of size $|D_i|=\varphi_i|\cup_j D_j|$. The residuals are initialized to zero $\Delta \W,\mathcal{R}_i,\mathcal{R}\leftarrow 0$. \\
\For{$t=1,..,T$}{
\For{$i \in I_t\subseteq \{1,..,\textnormal{[Number of Clients]}\}$ \textbf{in parallel}}{
\underline{Client $C_i$ does:}\\
\textbullet~ $\text{msg} \leftarrow \text{download}_{S \rightarrow C_i}(\text{msg})$\\
\textbullet~ ${\Delta \W} \leftarrow \text{decode}(\text{msg})$\\
\vspace{6pt}
\textbullet~ $\W_i \leftarrow \W_i + {\Delta \W}$ \\
\textbullet~ $\Delta \W_i \leftarrow \mathcal{R}_i+\text{SGD}(\W_i, D_i, b)-\W_i$ \\
\textbullet~ $\tilde{\Delta \W_i} \leftarrow \text{STC}_{p_{up}}(\Delta\W_i)$\\
\textbullet~ $\mathcal{R}_i \leftarrow \Delta \W_i - \tilde{\Delta \W_i}$ \\
\vspace{6pt}
\textbullet~ $\text{msg}_i \leftarrow \text{encode}(\tilde{\Delta \W_i})$\\
\textbullet~ $\text{upload}_{C_i \rightarrow S}(\text{msg}_i)$
}
\underline{Server $S$ does:}\\
\textbullet~ $\text{gather}_{C_i\rightarrow S}(\tilde{\Delta \W_i}),~i\in I_t$\\
\vspace{6pt}
\textbullet~ $\Delta \W \leftarrow \mathcal{R} + \frac{1}{|I_t|}\sum_{i\in I_t}\tilde{\Delta \W_i}$\\
\textbullet~ $\tilde{\Delta \W} \leftarrow \text{STC}_{p_{down}}(\Delta \W)$ \\
\textbullet~ $\mathcal{R} \leftarrow \Delta \W - \tilde{\Delta \W}$ \\
\textbullet~ $\W\leftarrow \W+\tilde{\Delta \W}$\\
\vspace{6pt}
\textbullet~ $\text{msg}\leftarrow \text{encode}(\tilde{\Delta \W})$\\
\textbullet~ $\text{broadcast}_{S\rightarrow C_i}(\text{msg}),~i=1,..,M$

}
\Return $\W$
\end{algorithm}

\section{Experiments}
\label{sec:experiments}
We evaluate our proposed communication protocol on four different learning tasks and compare it's performance to Federated Averaging and signSGD in a wide a variety of different Federated Learning environments. 

\textbf{Models and Datasets:} To cover a broad spectrum of learning problems we evaluate on differently sized convolutional and recurrent neural networks for the relevant Federated Learning tasks of image classification and speech recognition: 

\emph{VGG11* on CIFAR}: We train a modified version of the popular 11-layer VGG11 network \cite{simonyan2014very} on the CIFAR \cite{krizhevsky2014cifar} dataset. We simplify the VGG11 architecture by reducing the number of convolutional filters to [32, 64, 128, 128, 128, 128, 128, 128] in the respective convolutional layers and reducing the size of the hidden fully-connected layers to 128. We also remove all dropout layers and batch-normalization layers as regularization is no longer required. Batch-normalization has been observed to perform very poorly with both small batch sizes and non-iid data \cite{ioffe2017batch} and we don't want this effect to obscure the investigated behavior. The resulting VGG11* network still achieves 85.46\% accuracy on the validation set after 20000 iterations of training with a constant learning rate of 0.16 and contains 865482 parameters. 

\emph{CNN on KWS}: We train the four-layer convolutional neural network from \cite{konevcny2016federated} on the speech commands dataset \cite{warden2018speech}. The speech commands dataset consists of 51,088 different speech samples of specific keywords. There are 30 different keywords in total and every speech sample is of 1 second duration. Like \cite{zhao2018federated} we restrict us to the subset of 10 most common keywords. For every speech command we extract the mel spectrogram from the short time fourier transform, which results in a 32x32 feature map. The CNN architecture achieves 89.12\% accuracy after 10000 training iterations and has 876938 parameters in total.

\emph{LSTM on Fashion-MNIST}: We also train a LSTM network with 2 hidden layers of size 128 on the Fashion-MNIST dataset \cite{xiao2017fashion}. The Fashion-MNIST dataset contains 60000 train and 10000 validation greyscale images of 10 different fashion items. Every 28x28 image is treated as a sequence of 28 features of dimensionality 28 and fed as such in the the many-to-one LSTM network. After 20000 training iterations with a learning rate of 0.04 the LSTM model achieves 90.21\% accuracy on the validation set. The model contains 216330 parameters.

\emph{Logistic Regression on MNIST}: Finally we also train a simple logistic regression classifier on the MNIST \cite{lecun1998mnist} dataset. The MNIST dataset contains 60000 training and 10000 test greyscale images of handwritten digits of size 28x28. The trained logistic regression classifier achieves 92.31\% accuracy on the test set and contains 7850 parameters. 
 
The different learning tasks are summarized in Table \ref{tab:tasks}. 
In the following we will primarily discuss the results for VGG11* trained on CIFAR, however the described phenomena carry over to all other benchmarks and the supporting experimental results can be found in the appendix.

\begin{table}
\caption{Models and hyperparameters. The learning rate is kept constant throughout training.}
\label{tab:tasks}
\centering
\begin{tabular}{c|c|c|c|c}
\hline
Task & \makecell{VGG11* @\\CIFAR-10} & \makecell{CNN @\\KWS} & \makecell{LSTM@\\Fashion-MNIST} & \makecell{Logistic Reg.\\@ MNIST}\\
\hline
Iterations & 20000 & 10000 & 20000 & 5000\\
Learning Rate & 0.016 & 0.1 & 0.1 & 0.04\\
Momentum & 0.9 & 0.0 & 0.9 & 0.0\\
Base Accuracy & 85.46\% & 91.23\% & 90.21\% & 92.31\%\\
Parameters & 865482 & 876938 & 216330 & 7850\\
\hline

\end{tabular}
\end{table}

\begin{table}[t]
\caption{The base configuration of the Federated Learning environment in our experiments.}
\label{tab:parameters}
\centering
\begin{tabular}{c|c|c|c|c|c}
Parameter & \makecell{Number of\\Clients} & \makecell{Participation\\per Round} & \makecell{Classes\\per Client} & \makecell{Batch-\\Size} & \makecell{Balancedness}\\
\hline
\makecell{Value} & $N=100$ & $\eta=0.1$ & $c=10$ & $b=20$ & $\gamma=1.0$

\end{tabular}
\end{table}

\textbf{Compression Methods}:
We compare our proposed Sparse Ternary Compression method (STC) at a sparsity rate of $p=1/400$ with Federated Averaging at an "equivalent" delay period of $n=400$ iterations and signSGD with a coordinate-wise step-size of $\delta=0.0002$. At a sparsity rate of $p=1/400$ STC compresses updates both during upload and download by roughly a factor of $\times 1050$. A delay period of $n=400$ iterations for Federated Averaging results in a slightly smaller compression rate of $\times 400$. Further analysis on the effects of the sparsity rate $p$ and delay period $n$ on the convergence speed of STC and Federated Averaging can be found in Section \ref{sec:sparsity+delay} of the appendix. During our experiments, we keep all training related hyperparameters constant for the different compression methods. To be able to compare the different methods in a fair way, all methods are given the same budged of training iterations in the following experiments (one communication round of Federated Averaging uses up $n$ iterations, where $n$ is the number of local iterations).
%
%

\textbf{Learning Environment:} The Federated Learning environment described in Algorithm \ref{alg:DSGD} can be fully characterized by five parameters:
For the base configuration we set the number of clients to 100, the participation ratio to 10\% and the local batch size to 20 and assign every client an equally sized subset of the training data containing samples from 10 different classes. In the following experiments, if not explicitly signified otherwise, all hyperparameters will default to this base configuration summarized in Table \ref{tab:parameters}.  
We will use the short notations "Clients: $\eta N$/$N$" and "Classes: $c$" to refer to a setup of the Federated Learning environment in which a random subset of $\eta N$ out of a total of $N$ clients participates in every communication round and every client is holding data from exactly $c$ different classes.

\subsection{Momentum in Federated Optimization}
\begin{figure}[t]
\centering
\includegraphics[width=0.5\textwidth]{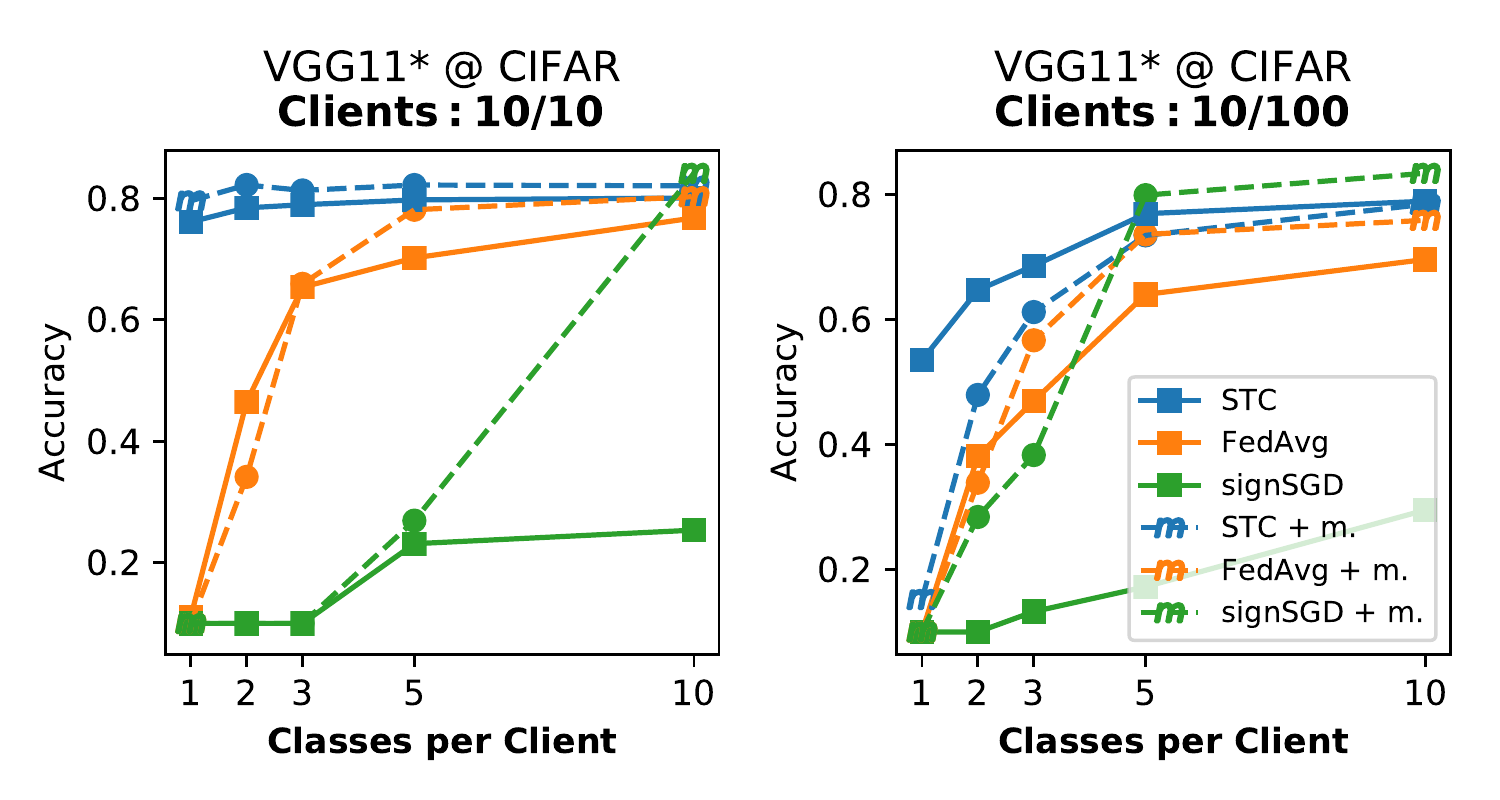}
\caption{Robustness of different compression methods to the non-iid-ness of client data on four different benchmarks. VGG11* trained on CIFAR. STC distinctively outperforms Federated Averaging on non-iid data. The learning environment is configured as described in Table \ref{tab:parameters}. Dashed lines signify that a momentum of $m=0.9$ was used.}
\label{fig:cpc}
\end{figure}

\begin{figure}[t]
\centering
\includegraphics[width=0.5\textwidth]{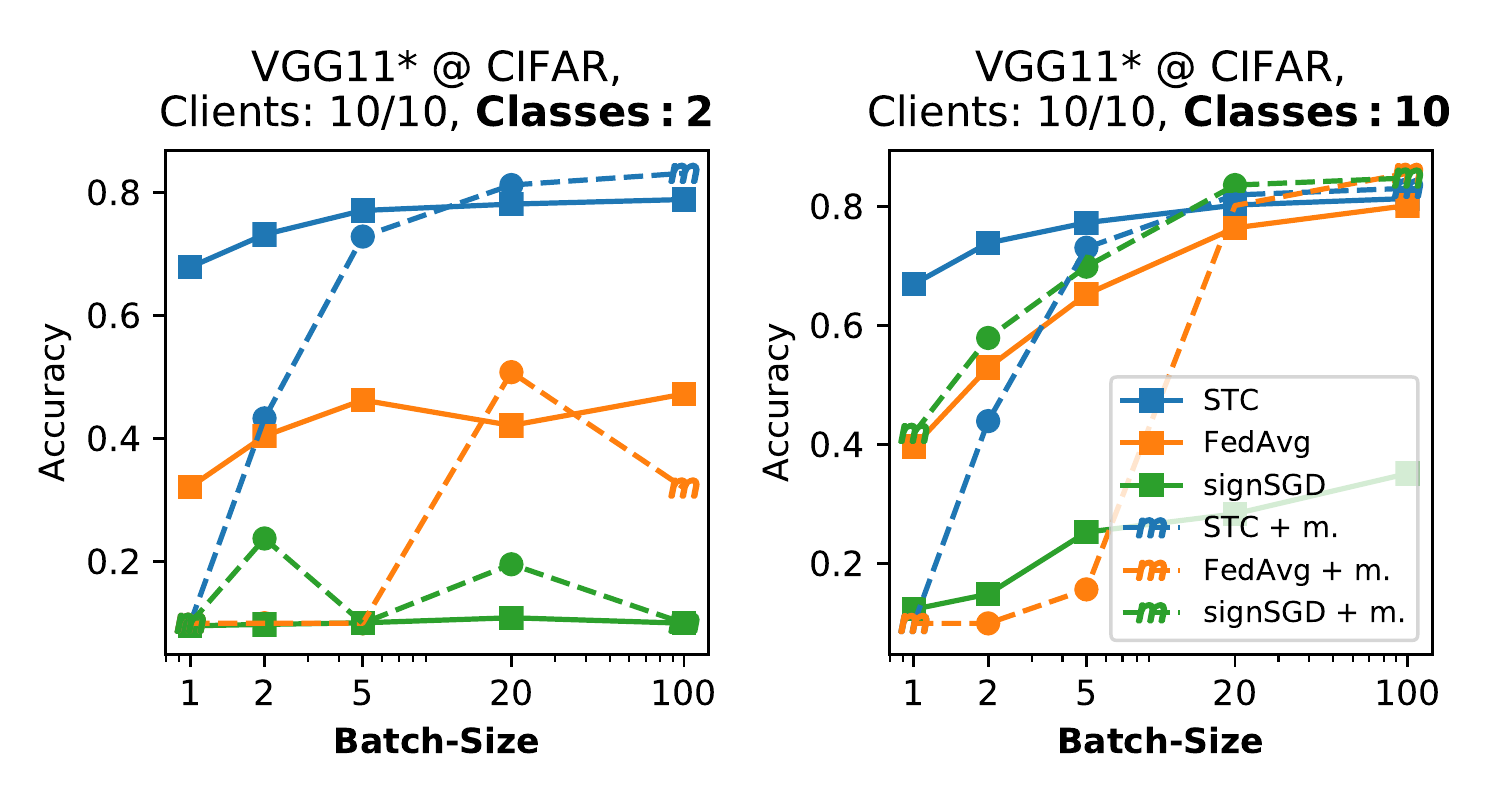}
\caption{Maximum accuracy achieved by the different compression methods when training VGG11* on CIFAR for 20000 iterations at varying batch sizes in a Federated Learning environment with 10 clients and full participation. In the left plot every client hold data from exactly two different classes, while in the right plot every client holds an iid subset of data.}
\label{fig:bs_cifar}
\end{figure}

We start out by investigating the effects of momentum optimization on the convergence behavior
of the different compression methods. Figures \ref{fig:cpc}, \ref{fig:bs_cifar}, \ref{fig:frac_cifar} and \ref{fig:unbalanced_cifar} show the final accuracy achieved by Federated Averaging ($n=400$), STC ($p=1/400$) and signSGD after 20000 training iterations in a variety of different Federated Learning environments. In all figures dashed lines refer to experiments where a momentum of $m=0.9$ was used during training, while solid lines signify that classical SGD was used. As we can see, momentum has significant influence on the convergence behavior of the different methods. While signSGD always performs distinctively better if momentum is turned on during the optimization, the picture is less clear for STC and Federated Averaging. We can make out three different parameters of the learning environment that determine whether momentum is beneficial or harmful to the performance of STC. If the participation rate is high and the batch size used during training is sufficiently large (Fig. \ref{fig:bs_cifar} left), momentum improves the performance of STC. Conversely, momentum will deteriorate the training performance in situations where training is carried out on small batches and with low client participation. The latter effect is increasingly strong if clients hold non-iid subsets of data (Fig. \ref{fig:cpc} right). 
These results are not surprising, as the issues with stale momentum described in \cite{lin2017deep} are enhanced in these situations. 
Similar relationships can be observed for Federated Averaging where again the size (Fig. \ref{fig:bs_cifar}) and the heterogeneity (Fig. \ref{fig:cpc}) of the local mini-batches determines whether momentum will have a positive effect on the training performance or not.     

When we compare Federated Averaging, signSGD and STC in the following, we will ignore whichever version of these methods (momentum "on" or "off") performs worse.

\subsection{Non-iid-ness of the Data}
Our preliminary experiments in Section \ref{sec:distributedtraining} have already demonstrated that the convergence behavior of both Federated Averaging and signSGD is very sensitive to the degree of iid-ness of the local client data, whereas sparse communication seems to be more robust. We will now investigate this behavior in some more detail. Figure \ref{fig:cpc} shows the maximum achieved generalization accuracy after a fixed number of iterations for VGG11* trained on CIFAR at different levels of non-iid-ness. Additional results on all other benchmarks can be found in Figure \ref{fig:appendix_cpc} in the appendix. Both at full (left plot) and partial (right plot) client participation, STC outperforms Federated Averaging across all levels of iid-ness. The most distinct difference can be observed in the non-iid regime, where the individual clients hold less than 5 different classes. Here STC (without momentum) outperforms both Federated Averaging and signSGD by a wide margin. In the extreme case where every client only holds data from exactly one class STC still achieves 79.5\% and 53.2\% accuracy at full and partial client participation respectively, while both Federated Averaging and signSGD fail to converge at all.

\subsection{Robustness to other Parameters of the Learning Environment}
\label{sec:learning_env}
\begin{figure}[]
\centering
\includegraphics[width=0.5\textwidth]{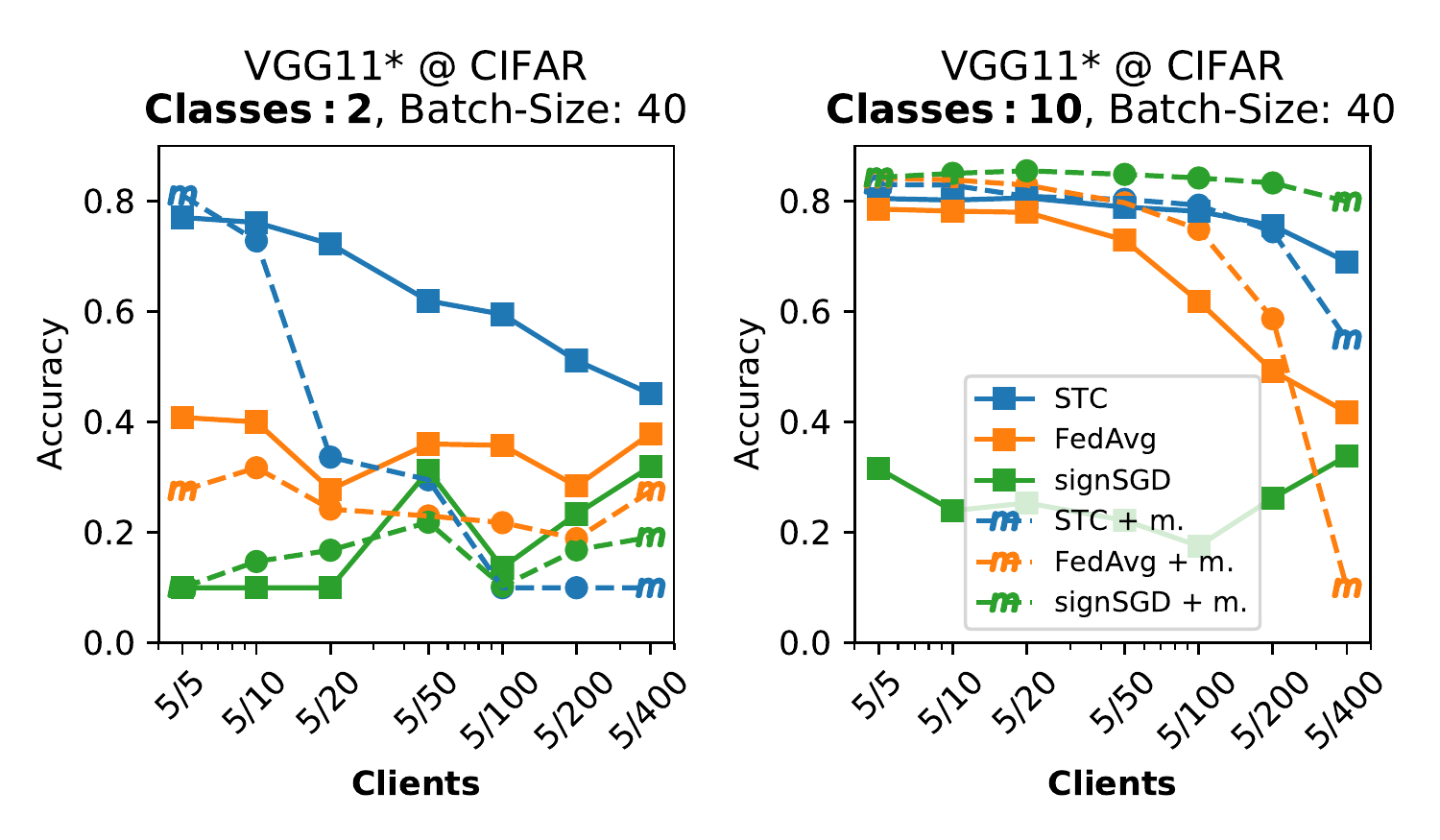}
\caption{Validation accuracy achieved by VGG11* on CIFAR after 20000 iterations of communication-efficient federated training with different compression methods. The relative client participation fraction is varied between 100\% (5/5) and 5\% (5/100). In the left plot every client hold data from exactly two different classes, while in the right plot every client holds an iid subset of data.}
\label{fig:frac_cifar}
\end{figure}

\begin{figure}[]
\centering
\includegraphics[width=0.25\textwidth]{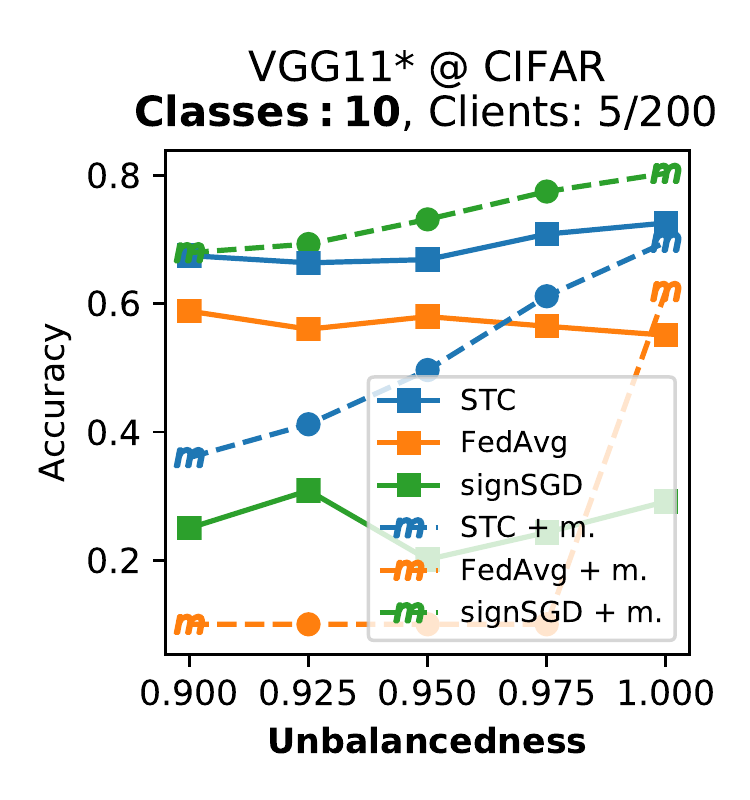}
\caption{Validation accuracy achieved by VGG11* on CIFAR after 20000 iterations of communication-efficient federated training with different compression methods. The training data is split among the client at different degrees of unbalancedness with $\gamma$ varying between 0.9 and 1.0.}
\label{fig:unbalanced_cifar}
\end{figure}

We will now proceed to investigate the effects of other parameters of the learning environment on the convergence behavior of the different compression methods. Figures \ref{fig:bs_cifar}, \ref{fig:frac_cifar}, \ref{fig:unbalanced_cifar} show the maximum achieved accuracy after training VGG11* on CIFAR for 20000 iterations in different Federated Learning environments. Additional results on the three other benchmarks can be found in Section \ref{sec:appendix_results} in the appendix. 

We observe that STC (without momentum) consistently dominates Federated Averaging on all benchmarks and learning environments.

\textbf{Local Batch Size:}
The memory capacity of mobile and IoT devices is typically very limited. As the memory footprint of SGD is proportional to the batch size used during training, clients might be restricted to train on small mini-batches only. 
Figure \ref{fig:bs_cifar} shows the influence of the local batch size on the performance of different communication-efficient Federated Learning techniques exemplary for VGG11* trained on CIFAR.
First of all, we notice that using momentum significantly slows down the convergence speed of both STC and Federated Averaging at batch sizes smaller than 20 independent of the distribution of data among the clients. 
As we can see, even if the training data is distributed among the clients in an iid manner (Fig. \ref{fig:bs_cifar} right) and all clients participate in every training iteration, Federated Averaging suffers considerably from small batch sizes. STC on the other hand demonstrates to be far more robust to this type of constraint. 
At an extreme batch size of 1 the model trained with STC still achieves an accuracy of 63.8\% while the Federated Averaging model only reaches 39.2\% after 20000 training iterations.

\textbf{Client Participation Fraction:}
Figure \ref{fig:frac_cifar} shows the convergence speed of VGG11* trained on CIFAR10 in a Federated Learning environment with different degrees of client participation. To isolate the effects of reduced participation, we keep the absolute number of participating clients and the local batch sizes at constant values of 5 and 40 respectively throughout all experiments and vary only the total number of clients (and thus the relative participation $\eta$). 
As we can see, reducing the participation rate has negative effects on both Federated Averaging and STC. The causes for  these negative effects however are different:
In Federated Averaging the participation rate is proportional to the effective amount of data that the training is conducted on in any individual communication round. If a non-representative subset of clients is selected to participate in a particular communication round of Federated Averaging, this can steer the optimization process away from the minimum and might even cause catastrophic forgetting \cite{goodfellow2013empirical} of previously learned concepts.
On the other hand, partial participation reduces the convergence speed of STC by causing the clients residuals to go out sync and increasing the gradient staleness \cite{lin2017deep}. The more rounds a client has to wait before it is selected to participate during training again, the more outdated it's accumulated gradients become.
We can observe this behavior for STC most strongly in the non-iid situation (Fig. \ref{fig:frac_cifar} left), where the accuracy steadily decreases with the participation rate. However even in the extreme case where only 5 out of 400 clients participate in every round of training STC still achieves a higher accuracy than Federated Averaging and signSGD.  
If the clients hold iid data (Fig. \ref{fig:frac_cifar} right), STC suffers much less from a reduced participation rate than Federated Averaging. If only 5 out of 400 clients participate in every round, STC (without momentum) still manages to achieve an accuracy of 68.2\% while Federated Averaging stagnates at 42.3\% accuracy. signSGD is affected the least by reduced participation which is unsurprising as only the absolute number of participating clients would have a direct influence on it's performance. Similar behavior can be observed on all other benchmarks, the results can be found in Figure \ref{fig:appendix_frac} in the appendix.
It is noteworthy that in Federated Learning it is usually possible for the server to exercise some control over the rate of client participation. For instance, it is typically possible to increase the participation ratio at the cost of a longer waiting time for all clients to finish. 

\begin{figure}
\centering
\includegraphics[width=0.5\textwidth]{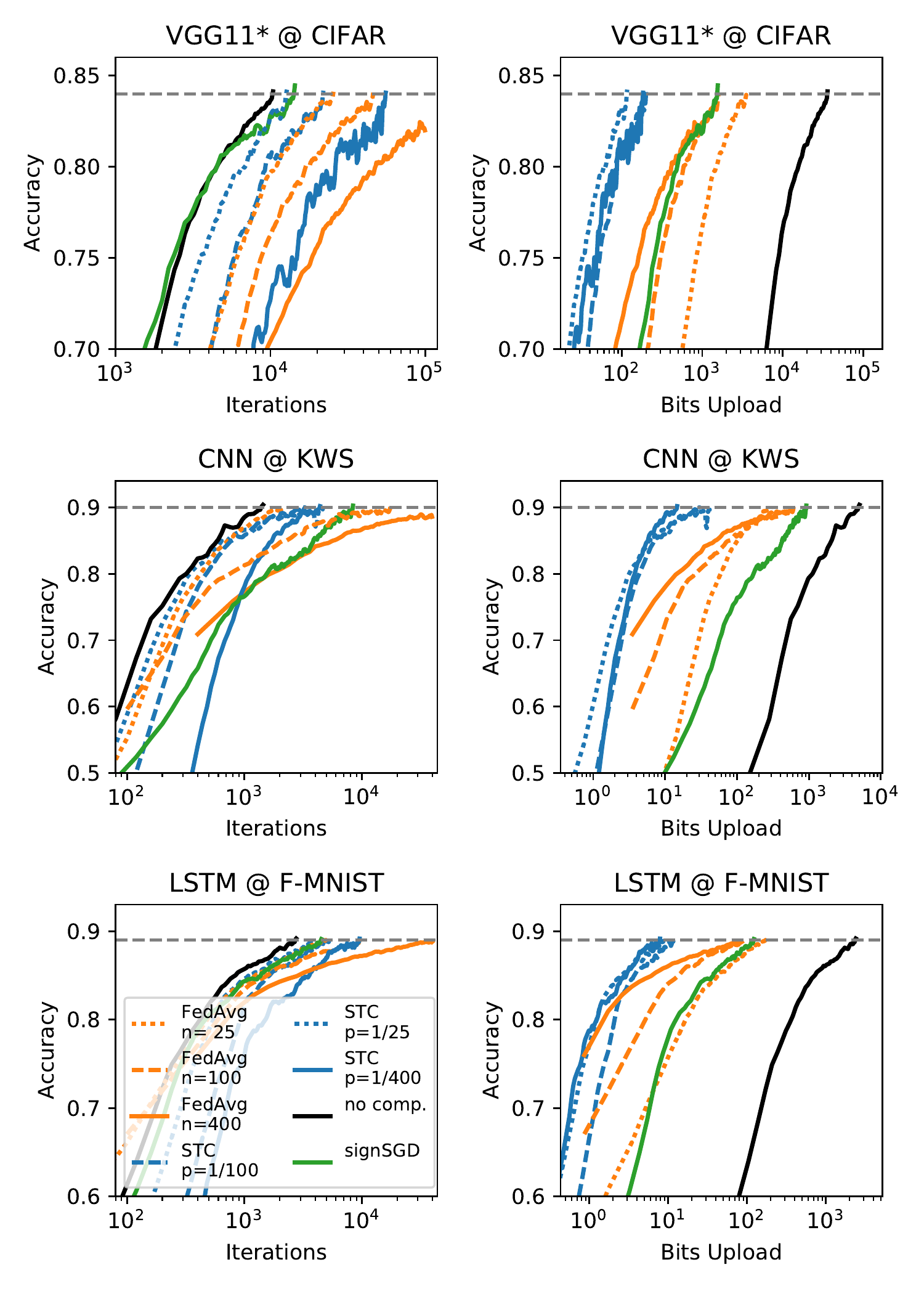}
\caption{Convergence speed of Federated Learning with compressed communication in terms of training iterations (left) and uploaded bits (right) on three different benchmarks (top to bottom) in an iid Federated Learning environment with 100 clients and 10\% participation fraction. For better readability the validation error curves are average-smoothed with a step-size of 5. On all benchmarks STC requires the least amount of bits to converge to the target accuracy.}
\label{fig:target_acc_big}
\end{figure}
\textbf{Unbalancedness}
Up until now, all experiments were performed with a balanced split of data in which every client was assigned the same amount of data points. In practice however, the datasets on different clients will typically vary heavily in size. 
To simulate different degrees of unbalancedness we split the data among the clients in a way such that the $i$-th out of $n$ clients is assigned a fraction 
\begin{align}
\varphi_i(\alpha, \gamma)=\frac{\alpha}{n}+(1-\alpha)\frac{\gamma^i}{\sum_{j=1}^n\gamma^j}
\end{align}
of the total data. The parameter $\alpha$ controls the minimum amount of data on every client, while the parameter $\gamma$ controls the concentration of data. We fix $\alpha=0.1$ and vary $\gamma$ between 0.9 and 1.0 in our experiments. To amplify the effects of unbalanced client data, we also set the client participation to a low value of only 5 out of 200 clients. Figure \ref{fig:unbalanced_cifar} shows the final accuracy achieved after 20000 iterations for different values of $\gamma$. Interestingly, the unbalancedness of the data does not seem to have a significant effect on the performance of either of the compression methods. Even if the data is highly concentrated on a few clients (as is the case for $\gamma=0.9$) all methods converge reliably and for Federated Averaging the accuracy even slightly goes down with increased balancedness. Apparently the rare participation of large clients can balance out several communication rounds with much smaller clients. These results also carry over to all other benchmarks (cf. Figure \ref{fig:appendix_unbalanced} in the appendix).  

\subsection{Communication-Efficiency}

Finally, we compare the different compression methods with respect to the number of iterations and communicated bits they require to achieve a certain target accuracy on a Federated Learning task.
As we saw in the previous Section, both Federated Averaging and signSGD perform considerably worse if clients hold non-iid data or use small batch sizes. 
To still have a meaningful comparison we therefore choose to evaluate this time on an iid environment where every client holds 10 different classes and uses a moderate batch size of 20 during training. This setup favors Federated Averaging and signSGD to the maximum degree possible!
All other parameters of the learning environment are set to the base configuration given in Table \ref{tab:parameters}. We train until the target accuracy is achieved or a maximum amount of iterations is exceeded and measure the amount of communicated bits both for upload and download. Figure \ref{fig:target_acc_big} shows the results for VGG11* trained on CIFAR, CNN trained on KWS and the LSTM model trained on Fashion-MNIST. We can see that even if all clients hold iid data STC still manages to achieve the desired target accuracy within a smallest communication budget out of all methods. STC also converges faster in terms of training iterations than the versions of Federated Averaging with comparable compression rate. Unsurprisingly we see that both for Federated Averaging and STC we face a trade-of between the number of training iterations ("computation") and the number of communicated bits ("communication"). On all investigated benchmarks however STC is pareto-superior to Federated Averaging in the sense for any fixed iteration complexity it achieves a lower (upload) communication complexity.  

Table \ref{tab:results} shows the amount of upstream- and downstream-communication required to achieve the target accuracy for the different methods in megabytes.
On the CIFAR learning task STC at a sparsity rate of $p=0.0025$ only communicates 183.9 MB worth of data, which is a reduction in communication by a factor of $\times 199.5$ as compared to the baseline with requires 36696 MB and Federated Averaging ($n=100$) which still requires 1606 MB. Federated Averaging with a delay period of 1000 steps does not achieve the target accuracy within the given iteration budget.

\begin{table}
\caption{Bits required for \emph{upload and/ download} to achieve a certain target accuracy on different learning tasks in an iid learning environment. A value of "n.a." in the table signifies that the method has not achieved the target accuracy within the iteration budget. The learning environment is configured as described in Table \ref{tab:parameters}.}
\centering
{\renewcommand{\arraystretch}{1.4}
\begin{tabular}{c|c|c|c}
 & \makecell{\textbf{VGG11*@CIFAR}\\\textbf{Acc. = 0.84}} & \makecell{\textbf{CNN@KWS}\\\textbf{Acc. = 0.9}} & \makecell{\textbf{LSTM@F-MNIST}\\\textbf{Acc. = 0.89}}\\
\hline
\hline
Baseline & \makecell{36696 MB /\\36696 MB} & \makecell{5191 MB /\\5191 MB}  &\makecell{2422 MB /\\2422 MB}   \\
\hline
\hline
signSGD & \makecell{1579.5 MB /\\6937.6 MB} & \makecell{925.17 MB /\\4063.6 MB} &  \makecell{123.31 MB /\\541.6 MB}  \\
\hline
\hline
\makecell{FedAvg $n=25$} & \makecell{3572.7 MB /\\3572.7 MB}  & \makecell{301.67 MB /\\301.67 MB}  & \makecell{174.79 MB /\\174.79 MB}  \\
\hline
\makecell{FedAvg $n=100$} & \makecell{1606.3 MB /\\1606.3 MB} &\makecell{617.3 MB /\\617.3 MB} & \makecell{83.94 MB /\\83.94 MB} \\
\hline
\makecell{FedAvg $n=400$} & n.a. & \makecell{350.78 MB /\\350.78 MB} & \makecell{86.53 MB /\\86.53 MB} \\
\hline
\hline
\makecell{STC $p=1/25$} & \makecell{\textbf{118.43 MB} /\\\textbf{1184.3 MB}}   & \makecell{43.57 MB /\\435.7 MB}   & \makecell{8.84 MB /\\88.4 MB}  \\
\hline
\makecell{STC $p=1/100$} & \makecell{202.2 MB /\\2022 MB}  & \makecell{31.0 MB /\\310 MB}  & \makecell{12.1 MB /\\121 MB} \\
\hline
\makecell{STC $p=1/400$} & \makecell{183.9 MB /\\1839 MB} & \makecell{\textbf{14.8 MB} /\\\textbf{148 MB}} & \makecell{\textbf{7.9 MB} /\\\textbf{79 MB}}\\
\hline
\hline
\end{tabular}
}

\label{tab:results}
\end{table}


\section{Lessons Learned}
\begin{figure}
\centering
\includegraphics[width=0.5\textwidth]{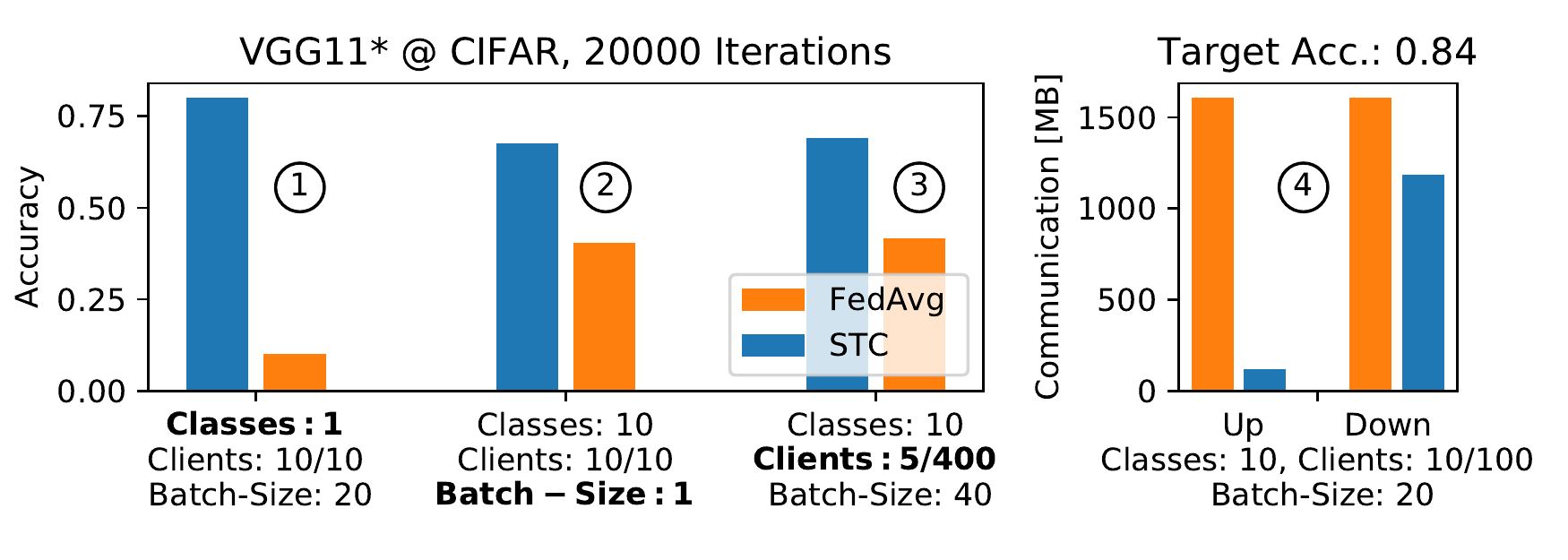}
\caption{Left: Accuracy achieved by VGG11* on CIFAR after 20000 iterations of Federated Training with Federated Averaging and Sparse Ternary Compression for three different configurations of the learning environment. Right: Upstream and downstream communication necessary to achieve a validation accuracy of 84\% with Federated Averaging and STC on the CIFAR benchmark under iid data and a moderate batch-size.}
\label{fig:summaryplot}
\end{figure}
We will now summarize the findings of this paper and give general suggestions on how to approach communication-constrained Federated Learning problems (cf. our summarizing Figure \ref{fig:summaryplot}): 
\begin{enumerate}
\item[\circled{1}] If clients hold non-iid data, sparse communication protocols such as STC distinctively outperform Federated Averaging across all Federated Learning environments (cf. Fig. \ref{fig:cpc}, Fig. \ref{fig:bs_cifar} left and Fig. \ref{fig:frac_cifar} left).
\item[\circled{2}] The same holds true if clients are forced to train on small mini-batches (e.g. because the hardware is memory constrained). In these situations STC outperforms Federated Averaging even if the client's data is iid (cf. Fig. \ref{fig:bs_cifar} right).
\item[\circled{3}] STC should also be preferred over Federated Averaging if the client participation rate is expected to be low as it converges more stable and quickly in both the iid and non-iid regime (cf. Fig. \ref{fig:frac_cifar} right).
\item[\circled{4}] STC is generally most advantageous in situations where the communication is bandwidth-constrained or costly (metered network, limited battery), as it does achieve a certain target accuracy within the minimum amount of communicated bits even on iid data (cf. Fig. \ref{fig:target_acc_big}, Tab. \ref{tab:results}).
\item[\circled{5}] Federated Averaging in return should be used if the communication is latency-constrained or if the client participation is expected to be very low (and \circled{1} - \circled{3} do not hold). 
\item[\circled{6}] Momentum optimization should be avoided in Federated Learning whenever either a) clients are training with small batch sizes or b) the client data is non-iid and the participation-rate is low (cf. Fig. \ref{fig:cpc}, \ref{fig:bs_cifar}, \ref{fig:frac_cifar}). 
\end{enumerate}

\section{Conclusion}
\label{sec:conclusion}
Federated Learning for mobile and IoT applications is a challenging task, as generally little to no control can be exerted over the properties of the learning environment.

In this work we demonstrated, that the convergence behavior of current methods for communication-efficient Federated Learning is very sensitive to these properties. On a variety of different datasets and model architectures we observe that the convergence speed of Federated Averaging drastically decreases in learning environments where the clients either hold non-iid subsets of data, are forced to train on small mini-batches or where only a small fraction of clients participates in every communication round. To address these issues we propose STC, a communication protocol which compresses both the upstream and downstream communication via sparsification, ternarization, error accumulation and optimal Golomb encoding. Our experiments show that STC is far more robust to the above mentioned peculiarities of the learning environment than Federated Averaging. Moreover, STC converges faster than Federated Averaging both with respect to the amount of training iterations and the amount of communicated bits, \emph{even} if the clients hold iid data and use moderate batch sizes during training.   

Our approach can be understood as an alternative paradigm for communication-efficient federated optimization which relies on high-frequent low-volume instead of low-frequent high-volume communication. As such it is particularly well suited for Federated Learning environments which are characterized by low latency and low bandwidth channels between clients and server.

\bibliographystyle{IEEEtran}
\bibliography{sample}

\appendices

\section{Encoding and Decoding}
\label{sec:coding}
To communicate the sparse ternary weight updates from clients to server and back from server to client we only need to transmit the positions of the non-zero elements in every tensor, along with exactly one bit to signify the sign ($\mu$ or $-\mu$). As the distances between the non-zero elements of the weight updates $\tilde{\Delta W}$ are approximately geometrically distributed for large layer sizes, we can efficiently encode them in an optimal way using the Golomb encoding \cite{golomb1966run}. The encoding scheme is given in Algorithm \ref{alg:encode}, while the decoding scheme is given in Algorithm \ref{alg:decode}.

\begin{algorithm} 
\caption{Golomb Position Encoding}\label{alg:encode}
\DontPrintSemicolon
\textbf{input:} sparse tensor $\Delta W^*$, sparsity $p$\\
\textbf{output:} binary message msg\\
\textbullet~ $\mathcal{I} \leftarrow \Delta W^*[:]_{\neq 0}$\\
\textbullet~ $\mathbf{b}^* \leftarrow 1+\lfloor \log_2(\frac{\log(\phi-1)}{\log(1-p)})\rfloor$\\
\For{$i=1,..,|\mathcal{I}|$}{
\textbullet~ $d \leftarrow \mathcal{I}_{i}-\mathcal{I}_{i-1}$\\
\textbullet~ $q \leftarrow  {(d-1)} \text{ div } {2^{\mathtt{b}^*}}$\\
\textbullet~ $r \leftarrow {(d-1)} \text{ mod } {2^{\mathtt{b}^*}}$\\
\textbullet~ msg.add($\underbrace{1, .., 1}_{q \text{ times}}$, 0, binary$_{\mathtt{b}^*}(r)$)\\
}
\Return msg
\end{algorithm}

\begin{algorithm} 
\caption{Golomb Position Decoding}\label{alg:decode}
\DontPrintSemicolon
\textbf{input:} binary message msg, bitsize $\mathbf{b}^*$, mean value $\mu$\\
\textbf{output:} sparse tensor $\Delta W^*$\\
\textbf{init:} $\Delta W^*\leftarrow 0\in \mathbb{R}^n$ \\
\textbullet~ $i\leftarrow0$; $q \leftarrow 0$; $j\leftarrow 0$\\
\While{$i<\text{size}(\msg)$}{
\uIf{$\msg[i]=0$}{
\textbullet~ $j\leftarrow j+q2^{\mathbf{b}^*}+\text{int}_{\mathbf{b}^*}(\msg[{i+1}],..,\msg[{i+\mathbf{b}^*}])+1$\\
\textbullet~ $\Delta W^*_j \leftarrow  \mu$\\
\textbullet~ $q\leftarrow 0$; $i\leftarrow i+\mathbf{b}^*+1$\\
}
\Else{
\textbullet~ $q\leftarrow q+1$; $i\leftarrow i+1$\\
}
}
\Return $\Delta W^*$
\end{algorithm}

\section{Data Splitting}
\label{sec:split_data}
We use the procedure described in Algorithm \ref{alg:split} to distribute the training data among the clients. In the resulting split every client holds a fixed proportion of the entire training data with $|D_i|=\varphi_i |D|$ and $|\{y:(x,y)\in D_i\}|=\textnormal{[Classes per Client]}$ $\forall
i=1,..,\textnormal{[Number of Clients]}$.

\begin{algorithm}
\caption{Data Spliting Strategy}\label{alg:split}
\DontPrintSemicolon
\textbf{input:} Data $D=\{(x_i, y_i), i=1,..,N\}$, Number of Clients $m$, Volume Distribution over Clients $\varphi=\{\varphi_1,..,\varphi_m\}$, [Classes per Client], [Number of different Classes]\\
\textbf{output:} Split $\mathcal{S}=\{D_1,..,D_m\}$\\
\textbf{init:} \\
\textbullet~ $D_i\leftarrow\{\}$, $i=1,..,m$\\
\textbullet~ Sort for classes: $A_j\leftarrow\{(x_i, y_i)\in D|y_i=j\}$, $j=1,..,\text{[Number of different Classes]}$\\
\For{$i=1,..,m$}{
\textbullet~ [Budget]$\leftarrow \varphi_iN$\\
\textbullet~ [Budget per Class]$ \leftarrow \text{[Budget]}/\text{[Classes per Client]}$\\
\textbullet~ $k\leftarrow \text{rand}(\{1,..,\text{[Number of different Classes]}\})$\\
\While{$\textnormal{[Budget]} > 0$}{
\textbullet~ $t\leftarrow\min\{\text{[Budget]}, \text{[Budget per Class]}, |A_k|\}$\\
\textbullet~ $\text{[Budget]}\leftarrow \text{[Budget]}-t$\\
\textbullet~ $B\leftarrow \text{randomSubset}(t,A_k)$\\
\textbullet~ $D_i \leftarrow D_i \cup B$\\
\textbullet~ $A_k\leftarrow A_k \backslash B$\\
\textbullet~ $k\leftarrow (k+1) \text{ mod } \textnormal{[Number of different Classes]}$\\

}
}

\end{algorithm}

\section{Combining Sparsity and Delay}
\label{sec:sparsity+delay}
\begin{figure}[H]
\centering
\includegraphics[width=0.5\textwidth]{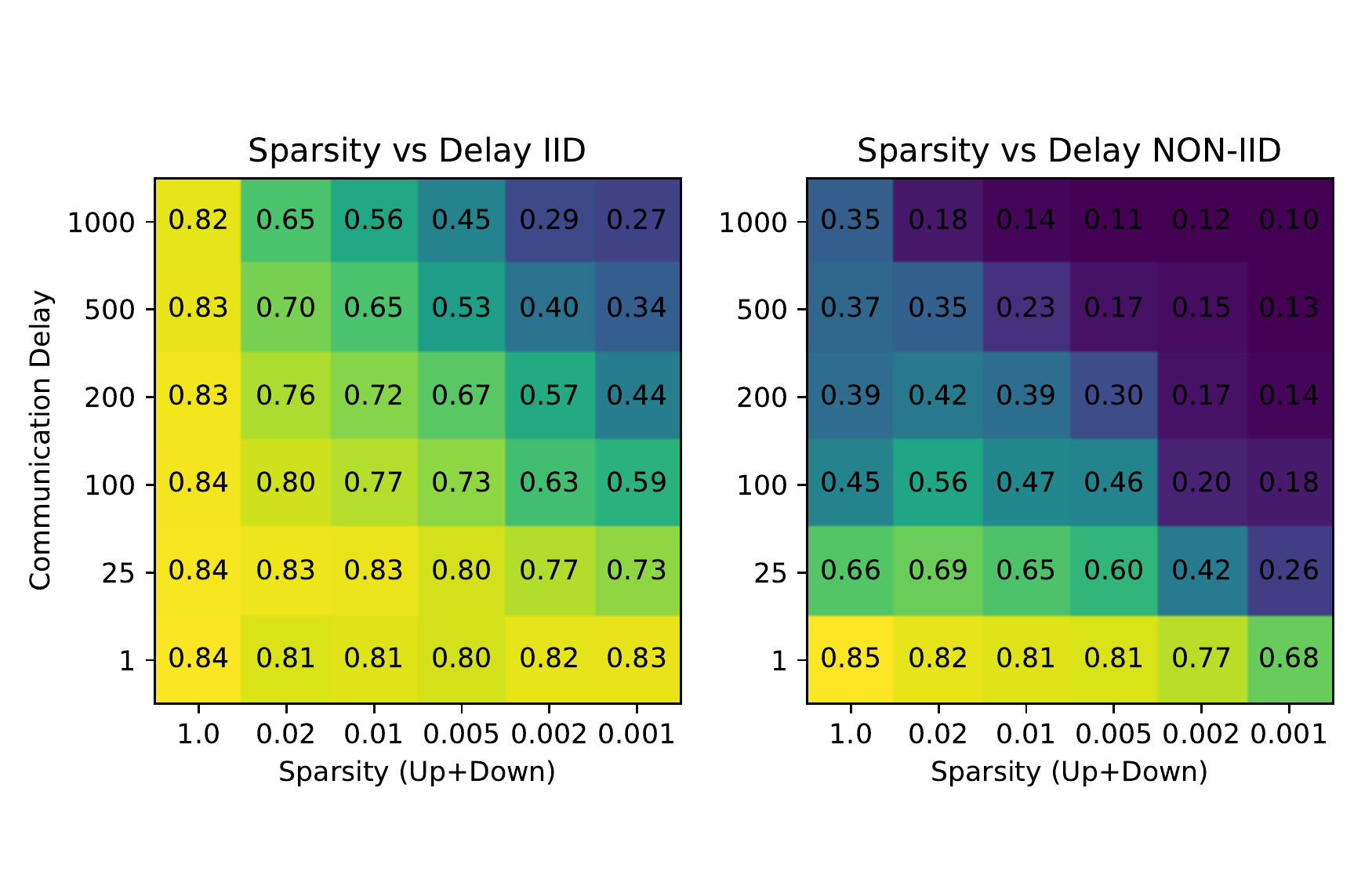}
\caption{Accuracy achieved after 10000 iterations for VGG11* trained on CIFAR with STC and Federated Averaging and combinations thereof at different rates of communication delay and sparsity in an iid (left) and non-iid (right) Federated Learning environment with 5 clients and full participation.}
\label{fig:sparse_delay}
\end{figure}

Figure \ref{fig:sparse_delay} compares accuracies achieved by Federated Averaging and STC at different rates of communication delay and sparsity respectively after 10000 iterations for VGG* trained on CIFAR. In the iid setting (left), sparsity and delay have a similar effect on the convergence speed. In the non-iid setting STC at any fixed sparsity rate always achieves higher accuracies than Federated Averaging at a comparable rate of communication delay. As already noted by \cite{sattler2018sparse} a combination of both techniques is possible and might be beneficial in situations where the communication is latency constraint.

\newpage
\section{Results: Learning Environments}
\label{sec:appendix_results}
\begin{figure}
\centering
\includegraphics[width=0.5\textwidth]{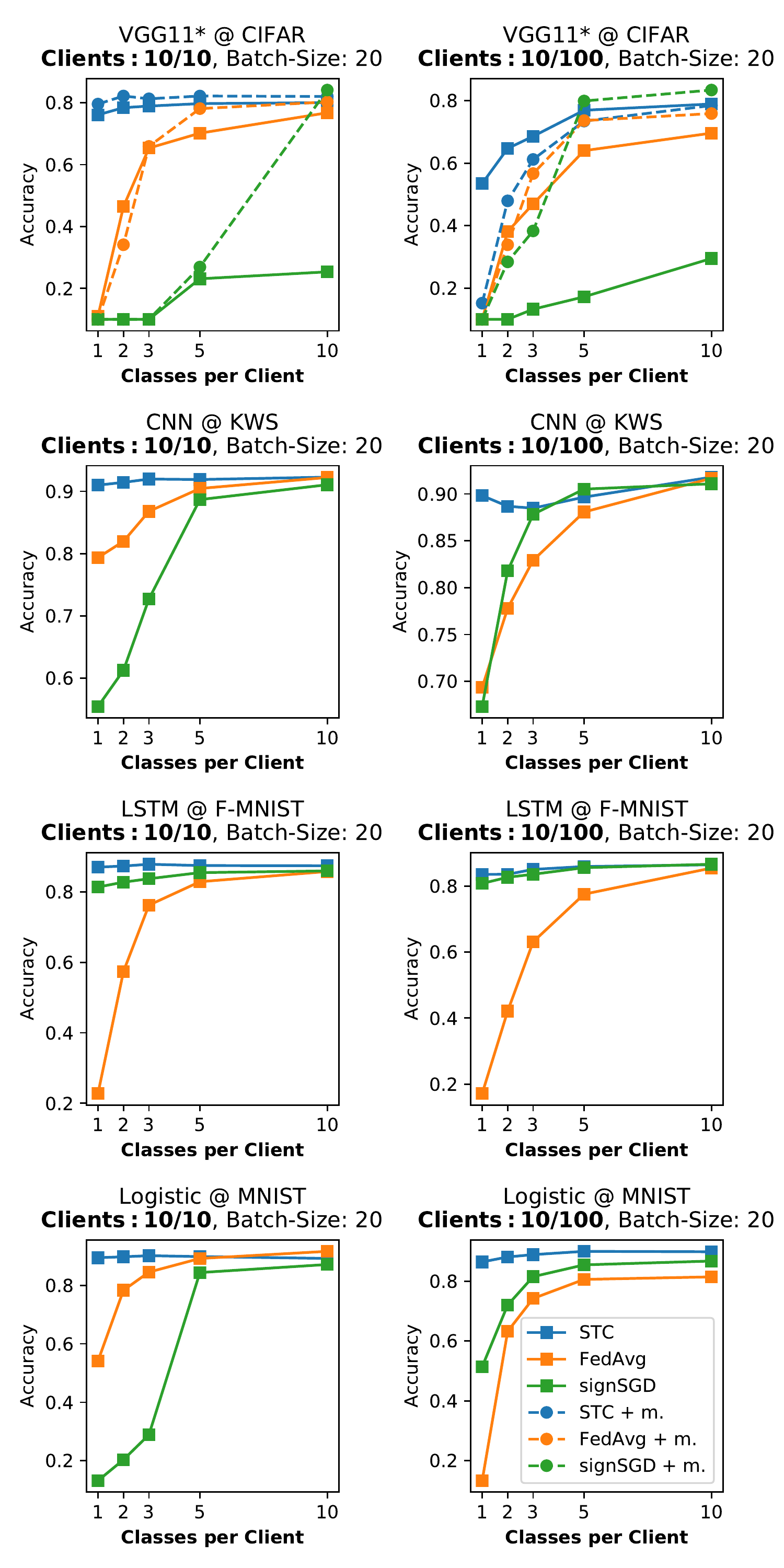}
\caption{Final accuracy achieved after training for a fixed number of iterations on four different learning tasks (top to bottom) and two different setups of the learning environment (left, right). Displayed is the relation between the final accuracy and the number of different classes in the clients datasets for different compression methods.}
\label{fig:appendix_cpc}
\end{figure}

\begin{figure}
\centering
\includegraphics[width=0.5\textwidth]{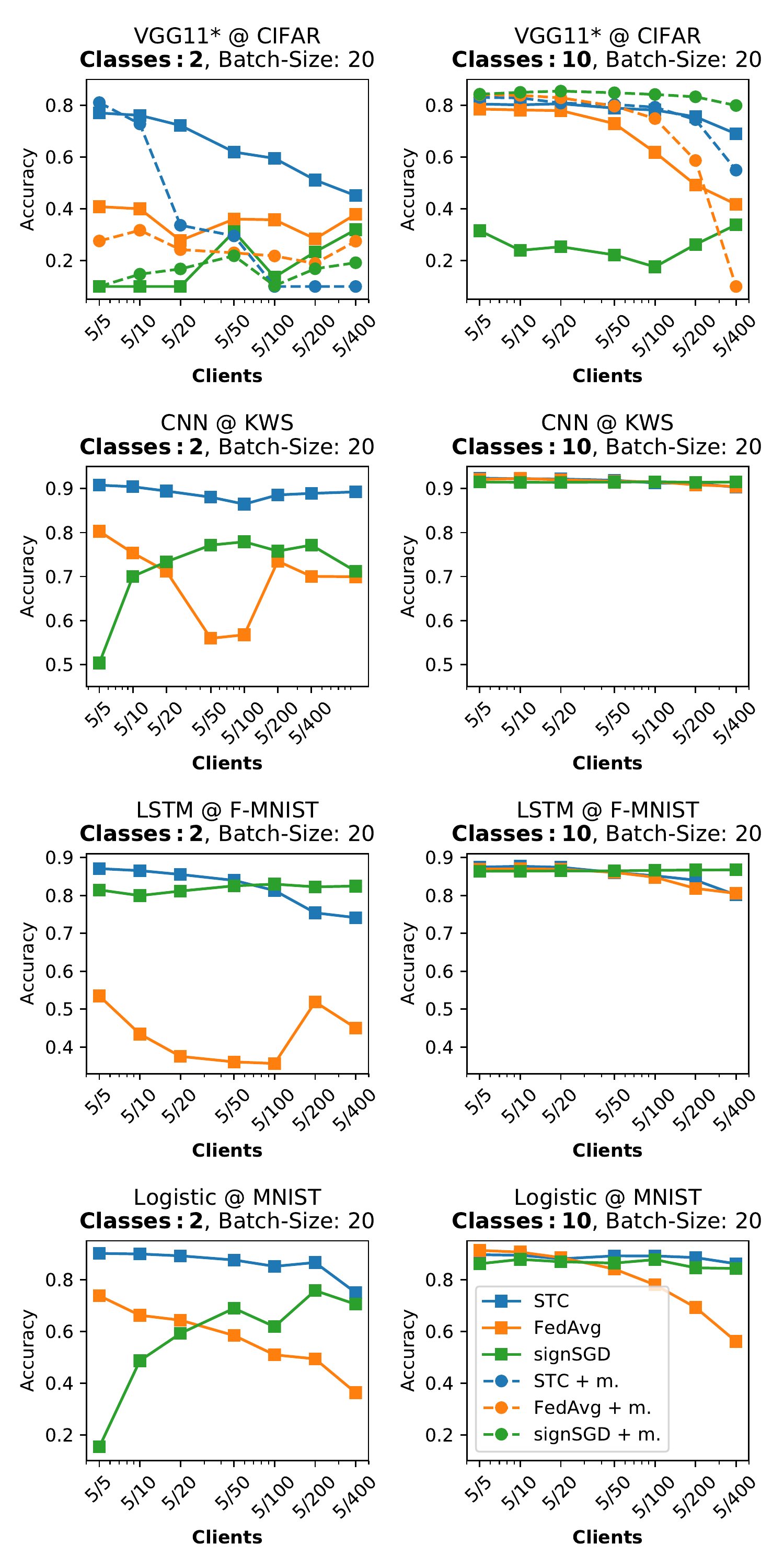}
\caption{Final accuracy achieved after training for a fixed number of iterations on four different learning tasks (top to bottom) and two different setups of the learning environment (left, right). Displayed is the relation between the final accuracy and the client participation fraction for different compression methods.}
\label{fig:appendix_frac}
\end{figure}

\begin{figure}
\centering
\includegraphics[width=0.5\textwidth]{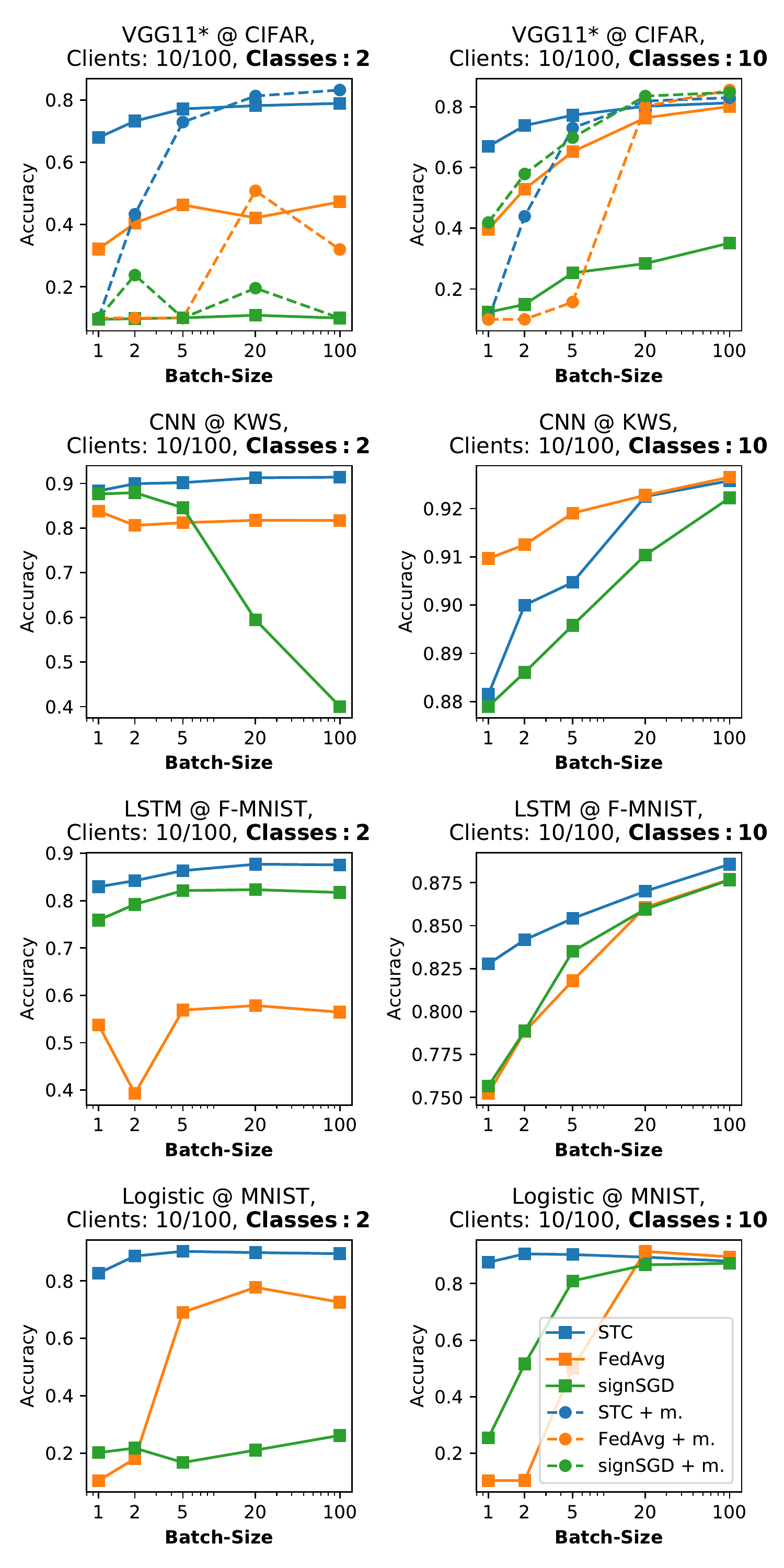}
\caption{Final accuracy achieved after training for a fixed number of iterations on four different learning tasks (top to bottom) and two different setups of the learning environment (left, right). Displayed is the relation between the final accuracy and the size of the mini-batches used during training for different compression methods.}
\label{fig:appendix_bs}
\end{figure}

\begin{figure}
\centering
\includegraphics[width=0.5\textwidth]{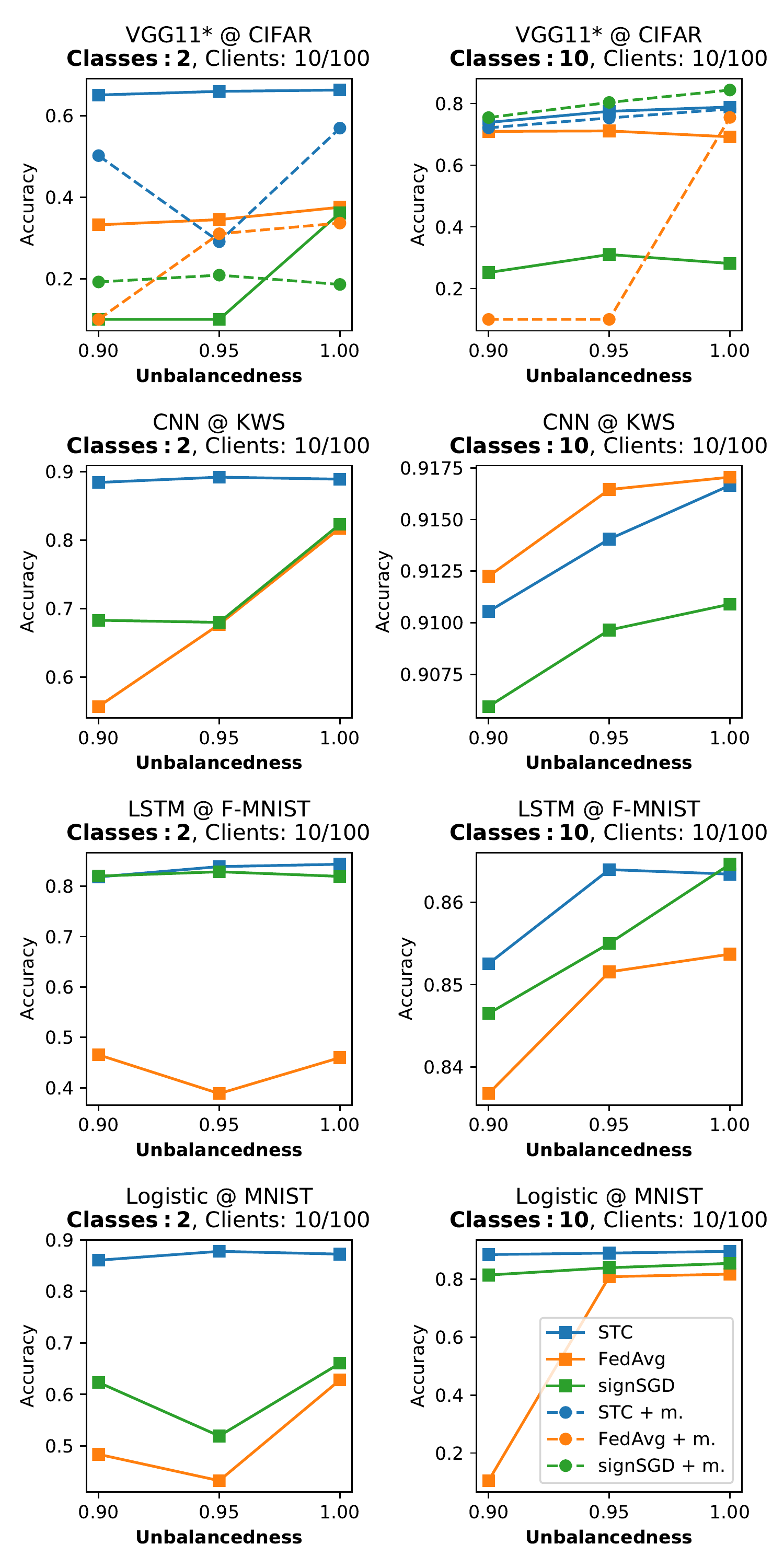}
\caption{Final accuracy achieved after training for a fixed number of iterations on four different learning tasks (top to bottom) and two different setups of the learning environment (left, right). Displayed is the relation between the final accuracy and the balancedness in size of the local datasets for different compression methods.}
\label{fig:appendix_unbalanced}
\end{figure}

Figures \ref{fig:appendix_cpc}, \ref{fig:appendix_frac}, \ref{fig:appendix_bs} and \ref{fig:appendix_unbalanced} show the final accuracy achieved by different compressed communication methods after a fixed number of training iterations on four different benchmarks (top to bottom) and different variations of the learning environment. We can see that the curves on the other benchmarks follow the same trends as the ones for the CIFAR benchmark.  

%
%
%
%

\newpage


\end{document}